%% file: main.tex
\pdfoutput=1

\documentclass[11pt]{article}

\usepackage[final]{acl}
\usepackage{times}
\usepackage{latexsym}

\usepackage[T1]{fontenc}
\usepackage[utf8]{inputenc}

\usepackage{microtype}
\usepackage{inconsolata}
\usepackage{graphicx}
\usepackage{multirow}
\usepackage{longtable}
\usepackage{booktabs}
\usepackage{xspace}

\usepackage{algorithm}
\usepackage{algpseudocode}
\usepackage{amssymb}
\usepackage{enumitem}

\input{math_commands}

\input{personal_commands}

\title{Efficient End-to-End Visual Document Understanding\\with Rationale Distillation}

\author{
\textbf{Wang Zhu}$^{\spadesuit\heartsuit}$\thanks{\quad Work done during internship at Google DeepMind}
\quad \textbf{Alekh Agarwal}$^{\clubsuit}$ 
\quad \textbf{Mandar Joshi}$^{\spadesuit}$ \\[4pt] 
\quad \textbf{Robin Jia}$^{\heartsuit}$ 
\quad \textbf{Jesse Thomason}$^{\heartsuit}$ 
\quad \textbf{Kristina Toutanova}$^{\spadesuit}$ \\[6pt]
$^\spadesuit$Google DeepMind
\quad $^\clubsuit$Google Research
\quad $^\heartsuit$University of Southern California \\[4pt]
}

\begin{document}
\maketitle
\begin{abstract}
Understanding visually situated language requires interpreting complex layouts of textual and visual elements. 
Pre-processing tools, such as optical character recognition (OCR), can map document image inputs to textual tokens, then large language models (LLMs) can reason over text.
However, such methods have high computational and engineering complexity. 
Can small pretrained image-to-text models accurately understand visual documents through similar recognition and reasoning steps instead?
We propose \ourmethodlong (\ourmethod), which incorporates the outputs of OCR tools, LLMs, and larger multimodal models as intermediate ``rationales'', and trains a small student model to predict both rationales and answers. 
On three visual document understanding benchmarks representing infographics, scanned documents, and figures, our \pts (282M parameters) student model finetuned with \ourmethod outperforms the base model by 4-5\% absolute accuracy with only 1\% higher computational cost.
\end{abstract}

\input{01_intro}
\input{02_task}
\input{03_method}
\input{04_exps}

\input{05_related}
\input{06_discussion}

\input{limit}
\input{acks}

\bibliography{custom}

\appendix
\clearpage

\input{appendix}

\end{document}

%% file: math_commands.tex
\usepackage{amsmath,amsfonts,bm}

\def\eqref#1{equation~\ref{#1}}

\def\1{\bm{1}}

\DeclareMathAlphabet{\mathsfit}{\encodingdefault}{\sfdefault}{m}{sl}
\SetMathAlphabet{\mathsfit}{bold}{\encodingdefault}{\sfdefault}{bx}{n}

\DeclareMathOperator*{\argmax}{arg\,max}

%% file: personal_commands.tex
\usepackage{pifont}
\newcommand{\cmark}{\ding{51}}%
\newcommand{\xmark}{\ding{55}}%

\newcommand{\infovqa}{InfoVQA\xspace}
\newcommand{\docvqa}{DocVQA\xspace}
\newcommand{\chartqa}{ChartQA\xspace}

\newcommand{\pts}{\textsc{Pix2Struct}\xspace}
\newcommand{\matcha}{\textsc{MatCha}\xspace}
\newcommand{\palm}{PaLM 2-L\xspace}
\newcommand{\palix}{PaLI-X\xspace}

\newcommand{\ourmethodlong}{Rationale Distillation\xspace} 
\newcommand{\ourmethod}{RD\xspace}
\newcommand{\qid}{QID\xspace}
\newcommand{\ansonly}{Ans-Only\xspace}

\newcommand{\taskone}{QRA\xspace}
\newcommand{\tasktwo}{ASR\xspace}
\newcommand{\taskthree}{QRACI\xspace}
\newcommand{\taskfour}{ALRCI\xspace}

\newcommand{\taskonelong}{Question, Rationale and Answer\xspace}
\newcommand{\tasktwolong}{Answer with Student Rationale\xspace}
\newcommand{\taskthreelong}{Question, Rationale and Answer on Cropped Images\xspace}
\newcommand{\taskfourlong}{Answer with Low-quality Rationale on Cropped Images\xspace}

%% file: 01_intro.tex
\begin{figure}[t]
    \centering
    \includegraphics[width=\linewidth]{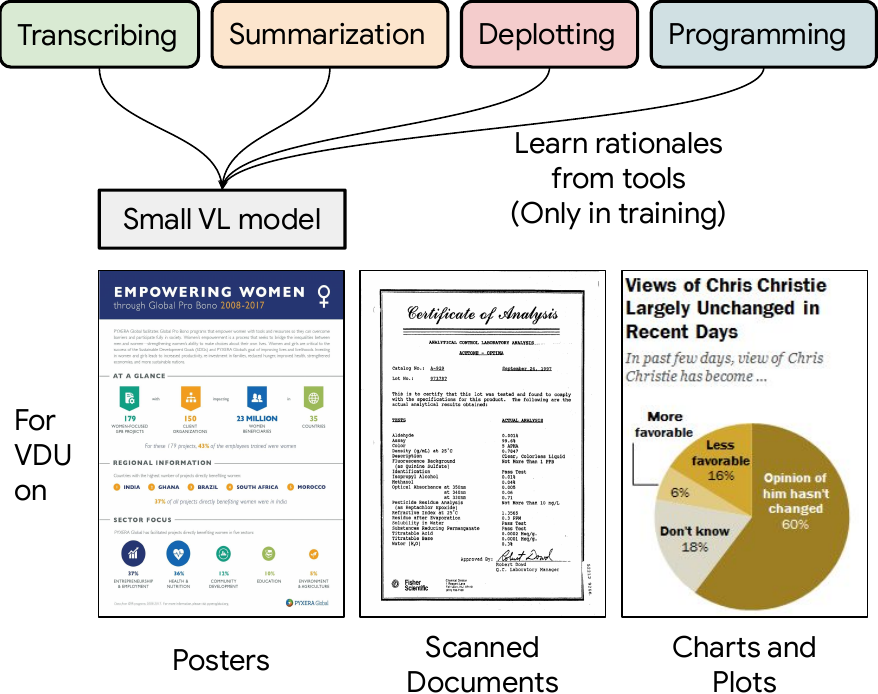}
    \caption{We synthesise the ability of recognizing and summarizing text, deplotting structured plots, and program generation into one small model and perform efficient rationale-based visual document understanding.}
    \label{fig:vdu_teaser}
\end{figure}

\begin{figure*}[ht]
    \centering
    \includegraphics[width=\textwidth]{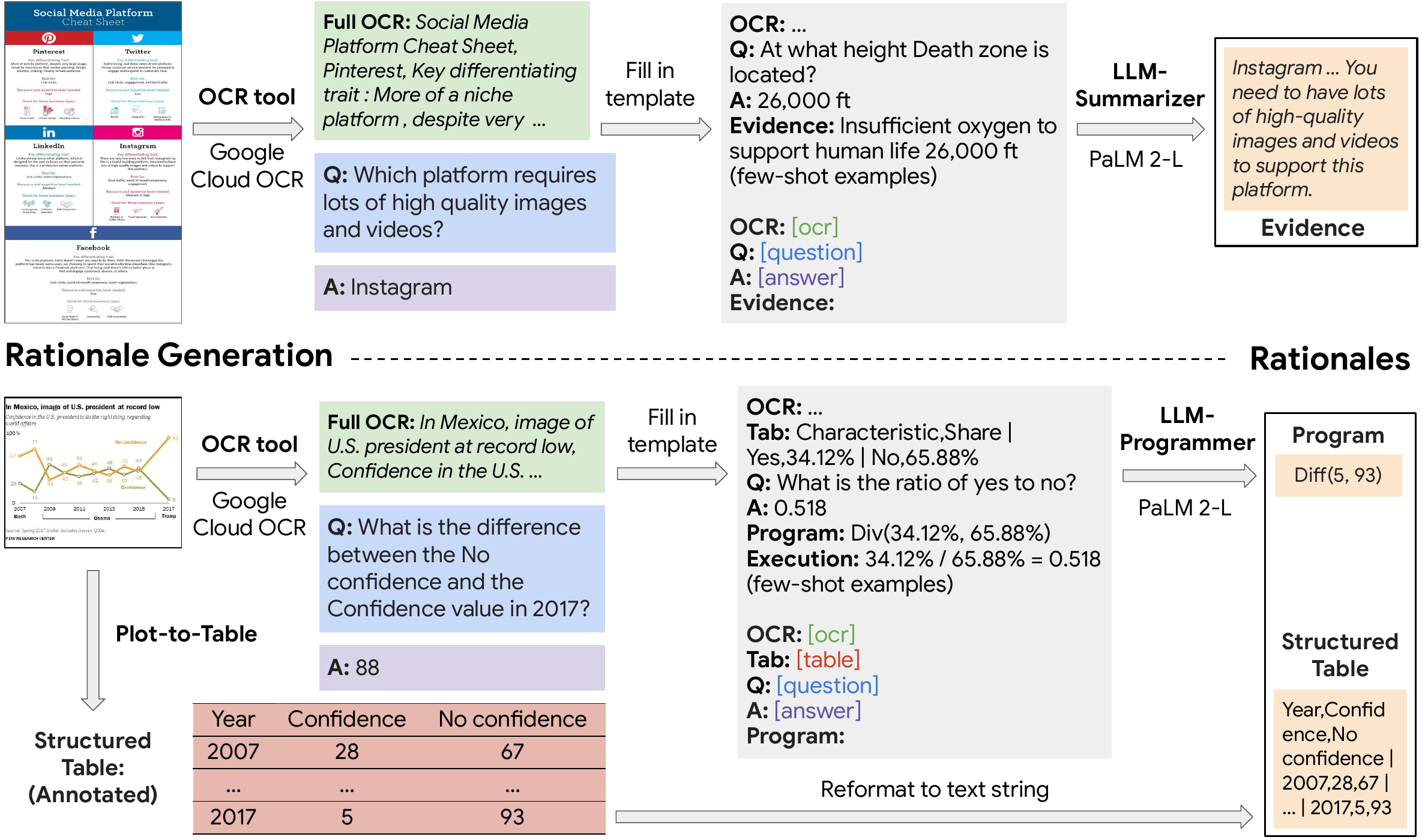}
    \caption{For training examples, we first generate the full OCR of each image with Google Cloud OCR. Depending on the dataset, we either use LLM-Summarizer (few-shot prompted \palm) to generate text evidence (top), or use  LLM-Programmer (also \palm) to generate a program based on both the OCR and available structured table source for the image (bottom).}
    \label{fig:rationale_generation}
\end{figure*}

\section{Introduction}

Information in the digital world is conveyed through text integrated with visual elements, such as complex layouts, figures, and illustrations. Answering user questions based on such visual documents requires models to recognize and connect text and layout to the user need.

While pretrained image-to-text multimodal models have demonstrated strong performance on visual document understanding (VDU) by directly mapping pixel-level input document images to answers corresponding to user queries~\cite{donut, pix2struct,chen2023palix,chen2023pali3}, state-of-the-art approaches benefit from the use of external tools.
Tools include OCR systems~\cite{chen2023palix, tilt, layoutlmv3}, structured table source extraction~\cite{liu-2022-deplot}, and LLMs reasoning over extracted information and the user query~\cite{liu-2022-deplot,perot2023lmdx}. Additional tools such as image captioning, object classification, and search engines have been used for other multimodal tasks~\cite{gpt4tools, zhang2023multimodal}.
However, the accuracy gains from these external components come at the cost of decreased computational efficiency and increased engineering complexity. 

In this work, we ask whether we can achieve high accuracy and efficiency by teaching a smaller model to learn from short rationales generated by external tools and expensive LLMs (see Figure~\ref{fig:vdu_teaser}).
We use a small student image-to-text model to perform VDU tasks by decomposing them into rationale prediction and answer steps, predicting the rationale and answer in sequence. 
The ``rationale'' can be any intermediate textual information that helps answer a question correctly:
for instance, it could be a subset of relevant text from the image as well as layout, structured information, and reasoning (see Figure~\ref{fig:rationale_generation}). 

The training data for VDU tasks of interest does not generally contain annotated ``rationales.'' 
It is also not known what types of sufficiently succinct rationales, even if available, would be useful for a small image-to-text model. 
We take inspiration from related works on chain-of-thought distillation~\cite{shridhar-etal-2023-distilling, zhang2023multimodal} for text and multimodal tasks, borrowing techniques and adding novel components to address the specific challenges within the visual document understanding domain. 
We use chains of tools at training time to derive short rationales representing salient subtasks of the problem---recognizing text and layout, and deriving programs to encode numerical reasoning. 
To increase the quantity and validity of example rationales, and the student's robustness to incorrect predictions, we design data augmentation schemes and DAGGER-style~\cite{pmlr-v15-ross11a} loss functions, which improves the student's ability to benefit from intermediate predictions.

Our method takes advantage of task decomposition and reasoning, but offers the following advantages over other tool-using models:

\begin{itemize}[nosep,noitemsep]
    \item No OCR or other external tools used during inference, reducing engineering complexity. 
    \item Only a short, query-dependent rationale is predicted versus longer structures typically extracted by external tools, saving computation. 
    \item Computation is increased by only about 1\% (in FLOPS) compared to models that predict the answer directly.
\end{itemize}

We conduct experiments on three VDU benchmarks: InfoVQA~\citep{mathew2022infographicvqa}, DocVQA~\citep{mathew2021docvqa}, and ChartQA~\citep{masry-etal-2022-chartqa}.
We show accuracy improvements over models that predict answers directly.
For models based on \pts-Base (282M parameters), improvements are 4.0 and 4.6 points in ANLS on InfographicVQA and DocVQA respectively, and 3.3 / 7.7 points in relaxed accuracy on ChartQA's augmented and human sets, with similar improvements for larger \pts models (1.3B parameters).

%% file: 02_task.tex
\begin{figure*}[t]
    \centering
    \includegraphics[width=\textwidth]{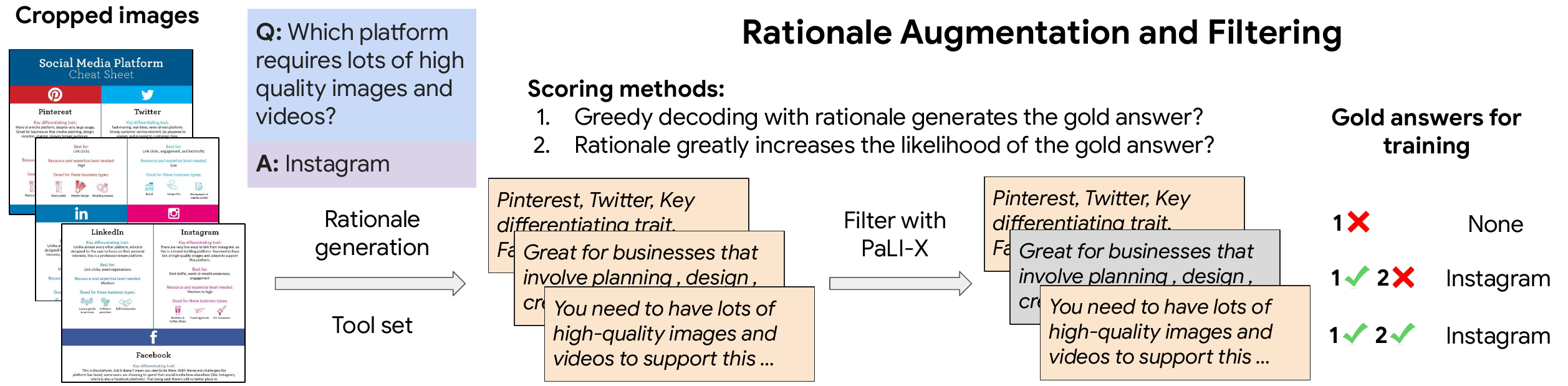}
    \caption{We first crop along the longer edge of the image to create multiple smaller square images. We generate rationales using the appropriate subset of tools (OCR, LLM-Summarizer, LLM-Programmer, Plot-to-Table) on these images, then categorize the examples and rationales with Multimodal-Verifier (\palix).}
    \label{fig:rationale_augmentation}
\end{figure*}

\section{Task definition}

In VDU, a model is given an image $I$ and user question $q$, and predicts text answer $a$. 
We focus on training a single small image-to-text model with parameters $\theta$ for this task.
Prior work in VDU trains such models by maximizing the training data log-likelihood according to an image-to-text (or image+text-to-text) model that directly generates the text answer $a$ given the input and makes predictions through greedy decoding, \textit{i.e.},  $\hat{a}=\argmax p_\theta(a \mid q, I)$ ~\cite{donut, pix2struct, chen2023palix, wang2022git}.

We assume that external tools such as OCR systems, LLMs, larger image-to-text models, or structured input image source information may be available at training time, but not inference time. We use such tools and metadata to derive rationales $r$ paired with training input-output examples $(I,q,a)$, and train a small student image-to-text model to predict rationales $r$ as an intermediate reasoning step, before predicting the answer $a$.

%% file: 03_method.tex
\section{\ourmethodlong}
We propose \ourmethodlong (\ourmethod), which distills rationales from a predefined set of tools, and trains a student model to predict the relevant rationales before predicting the answer.

\emph {Rationales} are sequences of text tokens of relevant information to arrive at the answer.
We consider two kinds of rationales $r$: natural language text evidence derived from the output of an OCR system; and tabular representation of charts in the input image concatenated with simple custom programs with predefined operations.
The tools we leverage are:
an OCR tool (Google Cloud OCR); 
Plot-to-Table, a converter that converts charts or plots to structured tables; 
a LLM-Summarizer (designed by us), which summarizes OCR text to evidences relevant to the question using a prompted \palm model~\cite{anil2023palm};
a LLM-Programmer (also designed by us and based on \palm), which generates simple programs for numerical reasoning tasks;
and a Multimodal-Verifier based on \palix~\cite{chen2023palix}, which verifies the quality of the rationales. 
We provide detailed descriptions of these tools in Appendix~\ref{appsec:impl}.
As the tools add heavy computation (for LLMs) or engineering complexity (for OCR), we depend on none of them at inference time.

In this section, we first discuss the process of generating the two types of rationales from tools (\S\ref{subsec:generation}). 
We then describe a data augmentation scheme for increasing the number of examples with rationales, and making student models more robust to potentially noisy rationales (\S\ref{subsec:augmentation}). 
Finally, we discuss training and inference for student models to predict the rationale and the answer (\S\ref{subsec:training}, \ref{subsec:inference}).

\subsection{Rationale generation from tools}
\label{subsec:generation}
\infovqa and \docvqa require a strong ability to recognize text, so we first use the OCR tool to extract the text from the image, then perform 5-shot prompting with LLM-Summarizer to generate question evidence (Figure~\ref{fig:rationale_generation}, top).
\chartqa focuses on numerical reasoning on charts, so we extract the full OCR text, obtain structured tables using Plot-to-Table,\footnote{We use provided \chartqa structured tables directly.} and then prompt LLM-Programmer (using 8 in-context examples) to generate a program to derive the answer. The concatenation of the structured table and the program are then used as the rationale (Figure~\ref{fig:rationale_generation}, bottom).
Detailed prompt templates are in Appendix~\ref{appsec:prompt}.

\subsection{Rationale augmentation and filtering}
\label{subsec:augmentation}
We aim to enable a small student model to reason over visual documents ranging over diverse formats and complexity. 
The expensive tools can typically generate high-quality rationales from such data, but it is a significant challenge for a small student model to match the quality of these rationales from a limited amount of training data. 
To overcome this challenge, we devise a data augmentation approach based on image cropping to greatly enlarge the number of examples available for rationale prediction, and to teach the model to use variable quality rationales in generating the final response.

\begin{algorithm}[ht]
  \caption{Rationale Filtering}
  \label{algo:filter}
\begin{algorithmic}[1]
\State {\bfseries Input:} image $I$; question $q$; answer $a$; rationale $r$; multimodal verifier with parameter $\phi$.
\State {\bfseries Output:} A tuple (the category of the rationale, the assigned answer used for training).
\If{$\argmax_{\hat{a}}p_{\phi}(\hat{a}\mid I,q,r)\neq a$}
    \State \textbf{return} ``irrelevant'', ``\texttt{None}''
\EndIf
\If{$[p_{\phi}(a\mid I,q,r)]^\lambda \geq p_{\phi}(a\mid I,q)$}
\State \textbf{return} ``useful'', $a$

\Else
\State \textbf{return} ``relevant but not useful'', $a$
\EndIf
\end{algorithmic}
\end{algorithm}

\paragraph{Cropping-based augmentation.} We crop the original image along the longer dimension, resulting in multiple square images (Figure~\ref{fig:rationale_augmentation}).
To minimize the possibility that the most relevant segment does not fit within any crop, we use a sliding window with adjacent croppings overlapping by half the image size (Algorithm~\ref{algo:aug}, appendix). 
For an input image $I$, we obtain $k$ cropped images $i_1, \dotsc, i_k$ and generate corresponding rationales for them as detailed above. As an example, in the InfoVQA dataset we observe an average of $k\approx4$.

\paragraph{Filtering relevant and useful examples.} While cropping significantly increases the size of our training dataset, many of the images might not contain information pertaining to the answer, and we may not be able to extract reasonable rationales. Including such examples in our dataset can amplify noise and make the problem more challenging for the student. So we carefully filter the augmented data to extract examples which are useful for rationale and/or answer prediction.
We use a powerful Multimodal-Verifier (\palix) with parameter $\phi$ to design two filters on VDU tasks (Algorithm 1).
We analyze the validity of \palix as a verifier model in Appendix~\ref{appsubsec:verify_pali}.

(1) The \emph{relevance filter} checks if the cropped image $i_j$ contains information for answering the question by comparing greedy decoding with the rationale as input against the gold answer:
${\argmax_{\hat{a}} p_{\phi}(\hat{a}\mid i_j,q,r_j)=a, j\in \{1..k\}}$ (row \#3 of Algorithm~\ref{algo:filter}).
For examples failing this filter, we replace the answer $a$ with \texttt{None} in the training data, assuming the cropped image is insufficient to generate the answer. 
For instance, the first cropped image of Figure~\ref{fig:rationale_augmentation} does not contain the gold answer ``Instagram'' and the example falls within the irrelevant category.
We still use the rationale $r_j$ for rationale prediction, since it could help distill the tool into the student model.

(2) The \emph{rationale filter} applies to examples that pass the relevance filter, and checks if the probability of the gold answer is sufficiently increased given the rationale (row \#6 of Algorithm~\ref{algo:filter}).
We use a factor $\lambda=2$ to avoid small perturbation caused by changing the format of the model prompt by concatenating the rationale. 
For examples that pass the relevance filter but not the rationale (row \#9), we regard the rationale $r_j$ as low-quality, and do not use it for learning rationale prediction.
For instance, the second cropped image of Figure~\ref{fig:rationale_augmentation} contains the gold answer ``Instagram'', but the tools do not generate a useful rationale.

We classify $(i_j, r_j)$ pairs into three categories (rows \#4,7,9 of Algorithm~\ref{algo:filter}), which determine their assigned answer $\bar{a}=a$ or \texttt{None} and the way their rationales are used in training.

\paragraph{Dataset balancing.} Most examples fail the relevance filter, and more than half of the ones that pass fail the rationale filter.
We subsample the examples with label \texttt{None} (row \#4) such that their number $n_{\text{row \#4}}$ $\leq$ $n_{\text{row \#9}} - n_{\text{row \#7}}$.

\input{tables/student_tasks}

\input{tables/pix2struct_main}

\subsection{Training student models}
\label{subsec:training}

In \ourmethodlong, we perform multi-task training for the student model, using tasks derived from the original and augmented data annotated with rationales. 
Tasks differ by their encoder and decoder inputs and decoder outputs (Table~\ref{tab:student_tasks}). 
We weight the four tasks equally (\textit{i.e.}, 0.25 for each), and train on a linear combination of them, with loss defined over the target output.

\paragraph{Distilling the tools directly.} This vanilla \taskonelong (\taskone) distillation setup teaches the model to take in the original image and predict $q$ (which can be read out from the image header), $r$ ( the intermediate rationale generated by the tools), and then by the answer $a$.

\paragraph{Robustifying against student rationale errors.} To help make the student model robust to its own mistakes, the \tasktwolong (\tasktwo) task provides question $q$ and student generated rationale $\hat{r}$ as decoder input for the student model to predict the answer. To generate such student rationales $\hat{r}$, we use a separately trained \pts-based student model, which learns to predict only rationales.

We sample three student generated rationales for each input example and use them as the low-quality rationales $\hat{r}$.  Since the training loss for \tasktwo is only applied to the answer prediction,  the \ourmethod student is not encouraged to replicate these noisy rationales, but to be able to recover from potential errors and predict the gold answer. We note that other than the difference of a separate student model generating the rationale, this is akin to student-forcing or DAGGER style approaches to structured prediction~\cite{pmlr-v15-ross11a}.

\paragraph{Leveraging cropped images.} In \taskthreelong (\taskthree), we use cropped images $i_j$ with rationales identified as useful (row \#7 of Algorithm~\ref{algo:filter}) or irrelevant (row \#4), to learn to predict those rationales and the original answer or \texttt{None}, respectively.
\taskfourlong (\taskfour) is similar to \tasktwo, taking cropped images as encoder input and providing low-quality rationales (row \#9) in the decoder input.

\subsection{Model architecture and inference details}
\label{subsec:inference}

\pts is an encoder-decoder model using a Transformer image encoder for an input image, and a Transformer-based decoder generating text.
Following~\citet{pix2struct}, we render the question $q$ as the header of the image $I$ for visual document understanding tasks and do not provide the question through a textual input channel.
We take \texttt{<s>} and \texttt{<answer>} as separators, and use the following encoding format for the decoder sequence:
{[\textit{question}] \texttt{<s>} [\textit{rationale}] \texttt{<answer>} [\textit{answer}]}.
As the decoder sequence length of \pts is 128 tokens,\footnote{Defined using \pts's tokenizer.} we trim the sequence before [\textit{answer}] to 108 tokens and leave 20 tokens for the answer.

If the rationale has programs, like in \chartqa, we put both the structured table and the program in the [\textit{rationale}] slot, using the format
{[\textit{rationale}] = [\textit{table}] \texttt{<program>} [\textit{program}]}.
As the structured table is usually long, we trim the sequence before [program] to 64, leaving 44 tokens for the program.

During inference, we evaluate only on the original, non-cropped images with greedy decoding.
To avoid generating answer \texttt{None}, we force the model to decode non-\texttt{None} after the \texttt{answer} token.

Note that student model's intermediate predictions are relatively short. 
The overall floating-point operations (FLOPs) compared to a baseline model that directly generates answers are increased by less than 1\% (see Appendix~\ref{appsec:flops} for a derivation).

%% file: tables/student_tasks.tex
\begin{table}[t]
\setlength{\aboverulesep}{2pt}
\setlength{\belowrulesep}{2pt}
\centering
\small
\tabcolsep 3pt
\begin{tabular}{lccc}
    Task name & Encoder input & Decoder input & Target output \\
    \toprule
    \taskone & $I$ & - & $q,r,a$ \\
    \tasktwo & $I$ & $q,\hat{r}$ & $a$ \\
    \taskthree & $i_j$ & - & $q,r_j,\bar{a}$ \\
    \taskfour & $i_j$ & $q,r_j$ & $\bar{a}$ \\
    \bottomrule        
\end{tabular}
\caption{We compute loss on the target output tokens for four student training tasks. Encoder input images have questions $q$ rendered as the header. Rationale $r$ (resp. $r_j$) is generated by tools on image $I$ (resp. $i_j$). Rationale $\hat{r}$ is generated by students.}
\label{tab:student_tasks}
\end{table}

%% file: tables/pix2struct_main.tex
\begin{table*}[t]
\setlength{\aboverulesep}{1pt}
\setlength{\belowrulesep}{1pt}
\centering
\small
\tabcolsep 8pt
\begin{tabular}{llrrrrrrrr}
    Model & Method & \multicolumn{4}{c}{Dev } & \multicolumn{4}{c}{Test } \\
    \cmidrule(lr){3-6}\cmidrule(lr){7-10}
    & & \infovqa & \docvqa & \multicolumn{2}{c}{\chartqa} & \infovqa & \docvqa & \multicolumn{2}{c}{\chartqa} \\
    & & & & aug. & human & & & aug. & human \\
    \toprule
    \multirow{3}{*}{Base} & \ansonly & 36.8 & 72.3 & 75.9 & 34.3 & 38.2 & 72.1 & 81.5 & 30.3 \\
    & \qid & 38.2 & 75.5 & 76.2 & 35.4 & 39.5 & 75.7 & 82.3 & 32.5 \\
    (282M) & \ourmethod (Ours) &  \bf 41.3 & \bf 76.3 & \bf 78.9 & \bf 36.7 & \bf 42.2 & \bf 76.7 & \bf 84.8 & \bf 38.0 \\
    \cmidrule(lr){2-10}
    & Oracle &  48.1 & 82.5 & 84.7 & 43.1 & - & - & - & - \\
    \midrule
    \multirow{3}{*}{Large} & \ansonly & 39.6 & 76.0 & 77.3 & 36.3 & 40.0 & 76.6 & 83.8 & 35.2 \\
    & \qid &  41.0 & 77.8 & 78.5 &  37.8 & 41.9 & 77.9 & 85.0 & 35.9 \\
    (1.3B) & \ourmethod (Ours) & \bf 43.5 & \bf 79.2 & \bf 81.6 & \bf 39.3 & \bf 44.3 & \bf 79.0 & \bf 88.6 & \bf 40.6 \\
    \cmidrule(lr){2-10}
    & Oracle & 53.5 & 84.0 & 85.8 & 46.5 & - & - & - & - \\
    \midrule
    ($\ge$5B) & SOTA & - & - & - & - & $^\dagger$62.4 & $^\dagger$88.6 & $^\ddagger$91.0 & $^\ddagger$67.6 \\
    \bottomrule
\end{tabular}
\caption{\pts-based results on three benchmarks. We show \ourmethodlong consistently outperforms the \ansonly and \qid baselines on both Base and Large models. Results marked by $^\dagger$ are from~\citet{chen2023pali3}, and ones marked by $^\ddagger$ are from~\citet{liu-2022-deplot}.}
\label{tab:pts_main}
\end{table*}

%% file: 04_exps.tex
\section{Experimental results}

We study the impact of rationale distillation across three benchmarks, analyze the contribution of each component of our approach, and the extent to which a single student model can match the capabilities of the external tools and LLMs it learns from.

\subsection{Dataset metrics}
\infovqa and \docvqa use the average normalized Levenshtein similarity (ANLS) score as the evaluation measure.
\chartqa uses relaxed accuracy (RA) and includes an easier augmented evaluation set and a harder human-generated evaluation set.

\subsection{Baselines}
\paragraph{\pts} We compare with the original \pts fine-tuning approach for both Base and Large models, where the model takes in an image $I$ with the question rendered as a header as encoder input and directly predicts $a$. 

\paragraph{\qid}
Fine-tuning tasks \taskone and \taskthree predict the question as part of the decoder output. To detect improvements due to reading out the question as an intermediate step, we compare to the question-in-decoder (\qid) setup, where the \pts model takes in $I$ in the encoder input and predicts the sequence $q,a$ separated by \texttt{<answer>}.

\paragraph{Oracle}
To establish an upper bound on performance of the student model if it was able to condition on the tool-generated high-quality rationales, we also compare to an oracle method on the development set.
We use the tool generated rationale $r$ during evaluation to get an oracle measure that uses information about the gold answer $a$.

We also describe other existing VDU approaches and compare \ourmethod to them in Table~\ref{tab:other_approaches} (Appendix~\ref{appsec:analysis}).

\subsection{Main results}
Table~\ref{tab:pts_main} evaluates our rationale distillation (RD) method against baselines.

\paragraph{Overall trends.}
 Overall, RD shows consistent improvements on \infovqa (4.0 and 4.3 points), \docvqa (4.6 and 2.4 points) and \chartqa-human (7.7 and 5.4 points) test sets  for both base and large model variants (respectively) over the \pts baseline. We also see that including the question in the decoder brings  benefits across all datasets and variants. Next, we discuss the value of rationale distillation in comparison to this stronger \qid baseline.

\paragraph{Textual rationales.}
Table~\ref{tab:pts_main} shows consistent improvements due to OCR and LLM-Summarizer rationales compared to the \qid baseline. RD records improvements of 2.7 and 2.4 points on \infovqa and 1.0 and 1.1 points on \docvqa for base and large variants respectively for the test set.

\paragraph{Table and program rationales.}
On \chartqa, we use rationales including Plot-to-Table (underlying tables for charts), as well as programs derived by LLM-Programmer (based on this table and OCR). Using such rationales results in improvements of 2.5 and 3.6 points respectively on base and large variants on the augmented set over the \qid baseline. We see even larger improvements: 5.3 and 4.7 points for base and large models, respectively, on the harder human eval set which requires more complex mathematical reasoning.

\paragraph{Accuracy and efficiency trade-off.}
We show that efficiency and accuracy can be improved at the same time.
The performance of the Base model with \ourmethod is better than that of the Large model with \ansonly; the inference FLOPs of the former ($\sim$2.65E+12) are also lower than those of the latter ($\sim$4.63E+12; Appendix~\ref{appsec:flops} shows a derivation).

On the other hand, \pts Large with \ourmethod still shows gaps compared to the SOTA methods --- PaLI-3 with OCR~\cite{chen2023pali3} on \infovqa and \docvqa, and a tool use case with deplotting and prompted LLM~\cite{liu-2022-deplot}.
It is worth noting that these methods use more than 10 times the FLOPs of the \pts Large model and also use more data.

\subsection{Analysis}
Using Base-sized models, we first explore two simple inference method variants that can improve model performance. We then ablate the impact of the different tasks designed to drive student model learning, and compare \ourmethod to additional baselines using cropping-based augmentations and tool-generated rationales in alternate ways. Finally, we analyze the quality of our student-generated rationales relative to ones predicted by external tools, and examine the types of questions that benefit most from \ourmethodlong.

\input{tables/voting}

\paragraph{Top-$n$ voting in inference.}
We can naturally apply top-$n$ voting during inference, which is similar to making predictions using self-consistency in chain-of-thought ~\cite{selfcons}. 
We simply perform beam search decoding with a beam size of $n=5$ and aggregate the probabilities of the distinct answers appearing in these hypotheses.
We choose the answer (that is not \texttt{None}) with the highest aggregate probability as the final prediction. From Table~\ref{tab:voting}, we see that this leads to small but consistent improvements across datasets, albeit at an increased computation cost. 

\paragraph{What if we use an external calculator on the generated programs?}
Using a calculator -- an additional but computationally inexpensive tool -- could further enhance the capabilities of our models. 
For valid programs generated by student models, we use a calculator to carry out computations dictated by the programs, and take the output of the calculator to replace  model output.
For invalid programs, we keep using the model generated answer prediction.
We observe that on \chartqa, the calculator use, when combined with voting, leads to further improvements of 0.4 RA on the augmented set and 3.3 RA on the human set.

\paragraph{How much does each of the tasks aid the student in predicting helpful rationales?}

In Table~\ref{tab:ablation_multitask}, we tease apart the contribution of each training task. First, we see that a model which uses standard supervised training with \taskone (\textit{i.e.}, predicting the question, rationale and answer) performs worse than the \qid baseline.
This result suggests that it is important to make the student model robust to its own errors and expose it to rationales with varying degrees of relevance to the question. 

Augmenting \taskone with \tasktwo (training with predicted rationales) results in a gain of about 1.9 points absolute (row \#3).
The additional image, rationale and answer examples obtained through image cropping and verifier categorization bring further improvements of 1.2 points (row \#6).

\input{tables/ablation_multitask}

\paragraph{Cropping-based augmentation is substantially less beneficial without rationales.}
As a traditional data augmentation method in vision, cropping may improve the accuracy of visual document understanding independent of the usage of intermediate rationales. We perform experiments to understand the gain from cropping in the baseline systems.
We compare cropping augmentation with and without rationales using \pts-base on \infovqa and show the gains with rationales are substantially higher. For setting without rationales, we study both the Ans-Only and QID setups.

Under the Ans-Only setup, we use one best (scored by pix2struct-infovqa-large) cropped image per example for augmentation, adding roughly the same amount of data as in the rationale augmentation setup. The performance with augmentation using cropped images is 37.1 on the \infovqa dev set, which indicates a small improvement of 0.3 over the analogous setting without augmentation (w/o augmentation Ans-Only is 36.8).

Under the QID setup, we generate augmentation examples using all crops of the input images ($\sim$4 cropped images per original image); we then form a training set which is a mixture of the original examples and the cropped image examples. 
We perform a grid search over downsampling ratios for the cropped examples at 2x, 4x (roughly the same amount of data as rationale augmentation), 8x, and 16x.
Results for QID without augmentation via cropping are 38.2, and the best results with augmentation was 38.4 with 8x downsampling. 
The results for all settings were: 37.5, 38.0, 38.4, and 38.3 for the 2x, 4x, 8x, and 16x downsampling configurations, respectively.

These findings suggest that cropping augmentation without rationales leads to a small improvement, and low-quality augmentation may lead to losses. The gains from augmentation for \ourmethod are substantially higher.

\paragraph{Directly distilling the tools via multi-task training is beneficial, albeit less effective than \ourmethod.}
Prior work has predicted teacher rationales as an auxiliary task in a multi-task training setting ~\cite{hsieh-etal-2023-distilling}, while predicting the answer directly during inference time. 
Here we evaluate whether multi-task training is helpful and whether predicting rationales at inference time brings additional value.
We experiment with multitask training on \infovqa for base-sized \pts models. 
Here we do not use data augmentation via cropping.  
The training tasks are to predict the answer, the question, and the text evidence derived from OCR and LLM-Summarizer. 
At inference, we predict the answer only.  
The tasks take the image (including the rendered question), denoted as i+q, and an instruction and produce corresponding outputs, as follows:  (1) given i+q and instruction “ans”, predict the answer; (2) given i+q and the instruction “ques”, predict the question; (3) given i+q and the instruction “ocr”, predict the text evidence. 
During inference, we use the instruction from task (1).

We experiment with combinations of tasks and sampling data proportions.
The ablation results in Table~\ref{tab:multi_distill} shows the effectiveness of distilling the tools with multitask training. The baseline that uses a single task predicting the answer obtains ANLS 36.8 (as shown in Table 2), and multi-task training improves performance to 38.7 for the best mixture, while not adding any inference-time cost. However performance when predicting rationales at inference time (RD) is higher, at 40.1 ANLS.

\input{tables/distilling_multitask}

\begin{figure}[ht]
    \centering
    \scalebox{.8}{
    \includegraphics[width=0.95\linewidth]{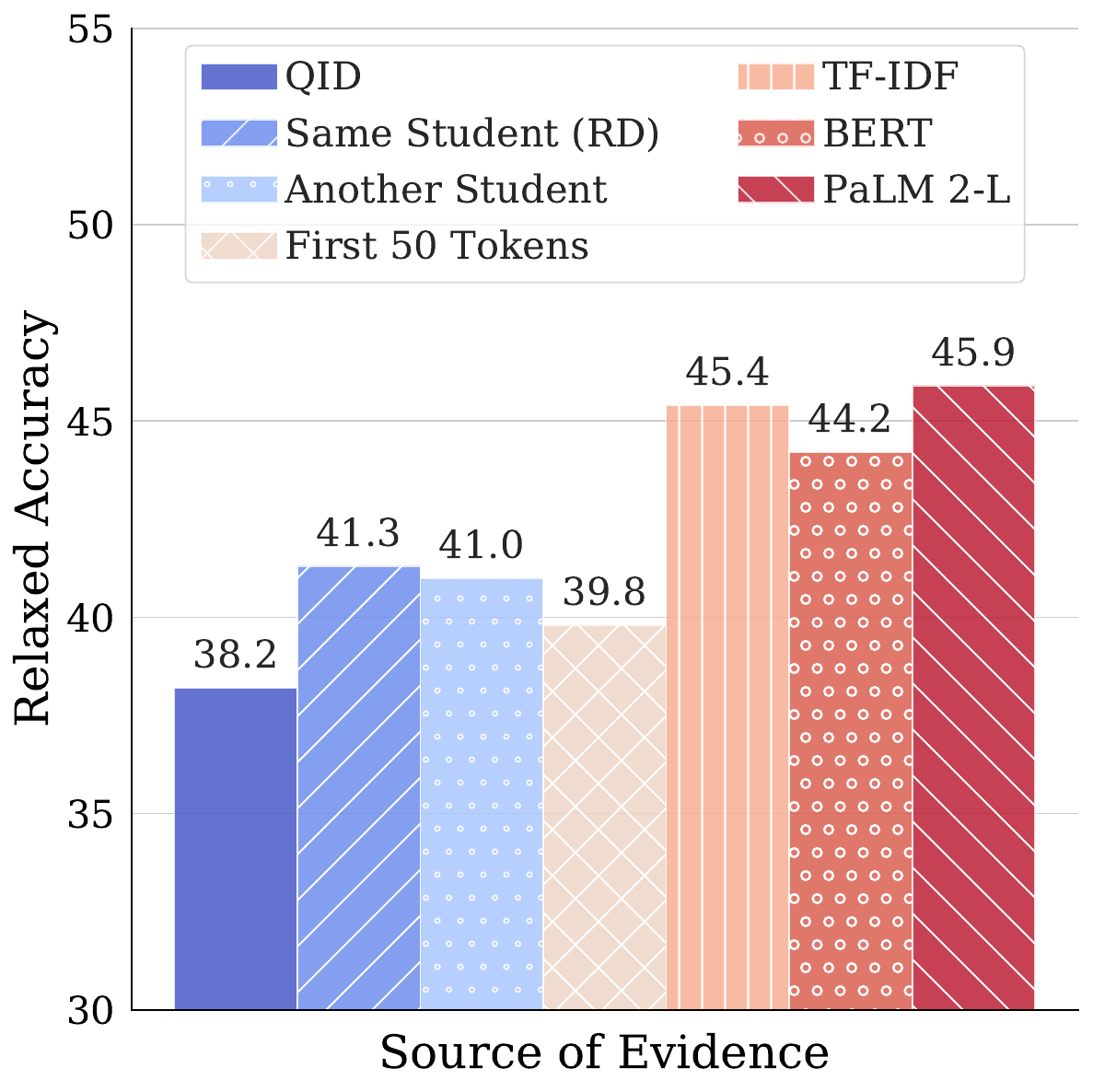}
    }
    \caption{We analyze the usefulness of student generated rationales. The systems are shown in order of increasing engineering complexity. All red bars use a  pipeline with Google Cloud OCR during inference. \ourmethod trades off between accuracy and complexity.}
    \label{fig:student_quality}
\end{figure}

\paragraph{What is the usefulness of the rationale generated by the student in comparison to external tools?}
On \infovqa, we analyze the usefulness of the student-generated rationale in comparison to evidence from the OCR tool and several ways to sub-select fragments of similar length from it including LLM-Summarizer without access to gold answer (based on \palm) (Figure~\ref{fig:student_quality}). 
The systems are shown (from left to right) in order of increasing computation costs and engineering complexity.
All methods except \qid are evaluated with \pts-Base trained with \ourmethod, using corresponding rationales as decoder input during inference.

\input{tables/breakdown}

Since OCR outputs can be very long, we experiment with different methods for selecting 50-token segments. The simplest variant truncates the OCR output to the first segment of 50 tokens which results in a small gain (1.6 points) over \qid. More complex methods which select segments based on TF-IDF (7.2 points) or BERT-embedding based (6 points) similarity to the question result in larger gains. 
Finally, the rightmost red bar shows the performance with rationales from few-shot prompted \palm. For this experiment, we modified the prompting template for \palm, to generate rationales from the OCR without being given the answer. Specifically, we ask \palm to predict the evidence first, and the answer next and use only the evidence (and not the answer) from \palm as student decoder input.
This variant performs the best with a 7.7 point improvement over \qid. Overall, these results indicate a significant room for improvement in rationale prediction for student models. We also see that an external OCR tool would still provide benefits at the cost of added computation by the OCR system and, since OCR is relatively efficient, the more significant cost of increased engineering complexity and potential service fees for production solutions.

\paragraph{Breakdown analysis of the improvement.}
The InfoVQA leaderboard provides a breakdown of model performance over subsets categorized by answer type, evidence type, and question operations.  
We compare the performance of \ansonly models (ANLS 38.2) and RD+Voting (ANLS 42.2) in Table~\ref{tab:breakdown_infovqa}.
We observe large improvements when answers are text spans in the image or in the question.
The former type indicates the helpfulness of the intermediate rationales; the latter suggests the helpfulness of decoding the question before answering.

We see a 7.1 point gain when the evidence comes from a table or list, 8 points when the evidence comes from text, which implies the student can extract better rationales from such parts of the images, in comparison to parts with more complex layouts such as figures and maps.

We did not use programs as rationales for \infovqa, and we do not see large improvement on arithmetic and counting questions. Using programs as parts of the rationales in this and other types of tasks is a promising direction for future work.

%% file: tables/voting.tex
\begin{table}[t]
\setlength{\aboverulesep}{1pt}
\setlength{\belowrulesep}{1pt}
\centering
\small
\begin{tabular}{lrrrr}
    Method & \multicolumn{4}{c}{Dev Set} \\
    \cmidrule(lr){2-5}
    & \infovqa & \docvqa & \multicolumn{2}{c}{\chartqa}\\
    & & & aug. & human \\
    \toprule
    \ourmethod & 41.3 & 76.3 & 78.9 & 36.7 \\
    RD+Voting & 41.7 & 76.6 & 79.4 & 37.0 \\
    \bottomrule        
\end{tabular}
\caption{\ourmethod on \pts-Base with voting during inference. Decoding with voting shows small but consistent improvements across datasets.}
\label{tab:voting}
\end{table}

%% file: tables/ablation_multitask.tex
\begin{table}[t]
\setlength{\aboverulesep}{1pt}
\setlength{\belowrulesep}{1pt}
\centering
\small
\tabcolsep 10pt
\begin{tabular}{clr}
    \# & Task & \infovqa Dev Set \\
    \toprule
    1 & \taskone  & 36.7 \\
    \midrule
    2 & \qid & 38.2 \\
    \midrule
    3 & \taskone and \tasktwo &  40.1 \\
    4 & \taskone, \tasktwo and \taskthree &  41.0 \\
    5 & \taskone, \tasktwo and \taskfour &  40.5 \\
    6 & All 4 tasks &  41.3 \\
    \bottomrule
\end{tabular}
\caption{We conduct ablation study of different student training task combinations on the \infovqa dev set: Question, Rationale, Answer (QRA), Answer with Provided Rationale (APR) and analoguous tasks on Cropped Images (CI). We show the importance of both training to predict the gold rationales and training to predict the answer based on the noisy rationales (row \# 3), as well as the usefulness of image cropping augmentation (row \# 6).}
\label{tab:ablation_multitask}
\end{table}

%% file: tables/distilling_multitask.tex
\begin{table}[t]
\setlength{\aboverulesep}{1pt}
\setlength{\belowrulesep}{1pt}
\centering
\small
\tabcolsep 12pt
\begin{tabular}{lcr}
    Train tasks & Weight per task & Dev ANLS \\
    \toprule
    ans/ques/ocr & 1:1:1 & 37.9 \\
    ans/ques/ocr & 4:1:1 & 38.4 \\
    ans/ques/ocr & 8:1:2 & 38.7 \\
    ans/ques & 2:1 & 37.0 \\
    ans/ocr & 2:1 & 38.2 \\
    \bottomrule
\end{tabular}
\caption{Distilling the OCR directly with multitask training on \infovqa is beneficial, but worse than \ourmethod.}
\label{tab:multi_distill}
\end{table}

%% file: tables/breakdown.tex
\begin{table*}[t]
\setlength{\aboverulesep}{1pt}
\setlength{\belowrulesep}{1pt}
\small
\centering
\tabcolsep 3pt
\begin{tabular}{lcccccccccccc}
    Method & \multicolumn{4}{c}{Answer type} & \multicolumn{5}{c}{Evidence} & \multicolumn{3}{c}{Operation} \\
    \cmidrule(lr){2-5}\cmidrule(lr){6-10}\cmidrule(lr){11-13}
    & Image & Question & Multiple & Non & Table/ & Textual & Visual & Figure & Map & Comparison & Arithmetic & Counting \\
    & span & span & spans & span & List & & object \\  
    \toprule
    \ansonly & 41.5 & 43.8 & 16.6 & 30.1 & 33.5 & 49.7 & 23.8 & 36.3 & 32.6 & 23.4 & 40.4 & \bf 18.9 \\
    RD+Voting & \bf 46.6 & \bf 46.7 & \bf 18.8 & \bf 30.4 & \bf 40.6 & \bf 57.7 & \bf 28.0 & \bf 37.9 & \bf 36.5 & \bf 28.1 & \bf 41.2 & 17.7 \\
    \bottomrule        
\end{tabular}
\caption{Breakdown ANLS score on different types of questions and answers from \infovqa test set. \ourmethod  benefits questions related to text or table evidence most.}
\label{tab:breakdown_infovqa}
\end{table*}

%% file: 05_related.tex
\section{Related work}
Using tools to augment the input in a prediction problem can be seen as using additional reasoning steps, \textit{i.e.}, calling a tool with a set of arguments and integrating its result with the rest of the context.
Much prior work on VDU has relied on calling OCR~\cite{tang2022unifying, appalaraju2021docformer, layoutlmv3}, object detector~\cite{kim2023cream}, or de-plotting tools~\cite{liu-2022-deplot}.
Such works have not attempted to recognize text or structured data as an intermediate reasoning step using the same small model.

On the other hand, the specific structure of reasoning chains through prompting LLMs has been shown to have significant impact~\cite{wei2023chainofthought, zhou2023leasttomost, khot2023decomposed, Yao2023TreeOT}.
Distilling these text rationales from large teacher models has been shown successful by chain-of-thought distillation works on NLP benchmarks~\cite{shridhar-etal-2023-distilling, li-etal-2023-symbolic} and ScienceQA~\cite{zhang2023multimodal, wang2023t}.
Toolformer~\cite{Schick2023ToolformerLM} trains smaller language models to call tools. 
Generic multimodal tool use solutions based on LLMs have also been proposed~\cite{gpt4tools}.
However, these works do not replicate the results of tool output and replace them for efficiency.

We marry the powerful ideas of taking intermediate reasoning steps from tools for accuracy, and distilling to small student models for efficiency, as we have proposed in \ourmethod.\footnote{We overview more related works in Appendix~\ref{appsec:related}.}

%% file: 06_discussion.tex
\section{Conclusions}

We showed that the visual document understanding ability of small image-to-text models can be improved by our proposed \ourmethodlong.
In \ourmethod, we obtain rationales for training examples using external tools and LLMs, and train small end-to-end student models to predict rationales as intermediate reasoning steps. We demonstrated the importance of designing student training tasks that make the model robust to irrelevant rationales.

\ourmethod leads to substantial improvements via textual evidence distillation on the text-heavy \infovqa \& \docvqa datasets, and via Plot-to-Table and program distillation on the numerical reasoning-focused \chartqa dataset.
Analysis shows the gains transfer to stronger models such as \matcha (Appendix B) and larger \pts models. Marginalizing over rationales and using a cheap calculator tool at inference time bring additional consistent benefits.
Controlled experiments show that \ourmethod offers a tradeoff between performance and computational cost/engineering complexity, in comparison to systems relying on tool pipelines.

%% file: limit.tex
\section*{Limitations}

Our study shows \ourmethod can teach small models to successfully generate and utilize two types of rationales: summarized OCR evidence, and structured table concatenated with a simple program.
A broader set of tools, such as object detection, image segmentation and captioning tools, can be further explored as rationales to enhance the ability of visual document understanding.

To use resources sparingly, we evaluate on the \pts series of models up to a size of 1.3B parameters (including the stronger \matcha model; see Appendix~\ref{appsec:analysis}).
In the future, \ourmethod could also be evaluated on other more powerful pretrained models for visual document understanding, such as PaLI-3 ~\cite{chen2023pali3} or ERNIE-Layout~\cite{peng-etal-2022-ernie}.

We focus on single-page visual document understanding, and have not explored the potential of \ourmethod on multi-page images. Multi-page image problems may have longer-distance dependencies, and require student models to generate more complex rationales as the intermediate reasoning steps.

We inherit the ethical concerns of existing LLMs and multimodal models, such as privacy considerations and potential misuse.
Here we use public peer-reviewed datasets to evaluate our method.
For use in deployed applications, the data for \ourmethod should be constructed with careful data curation. Privacy-sensitive documents which contain personal information, should be excluded from the training data to prevent potential privacy breaches and unintended consequences.

%% file: acks.tex
\section*{Acknowledgements}

We are grateful to Adam Fisch, Kenton Lee, Peter Shaw,  Yi Luan, Julian Eisenschlos,  Ming-Wei Chang, Fangyu Liu, Hexiang Hu, and other members of the Google research community for helpful discussion and feedback on this work.

%% file: appendix.tex
\section{Implementation details}
\label{appsec:impl}

\subsection{Description of used tools}
\paragraph{OCR} For all datasets, we begin with calling an off-the-shelf external OCR tool (Google Cloud OCR), which takes the image as input and outputs the full text recognized in the image together with location information (see Figure~\ref{fig:rationale_generation}).

\paragraph{LLM-Summarizer} OCR outputs can be quite long for some images, and not all text in the input image is directly relevant to a given question. To minimize computation spent on intermediate rationale prediction steps, we employ another powerful tool --- a prompted large language model \palm~\cite{anil2023palm}, to generate a significantly shorter span of text (less than 100 tokens), given the question, answer, and the full image OCR text (see Figure~\ref{fig:rationale_generation} top for an example). We sample a single evidence with temperature of 0.1 to obtain these rationales from \palm. 

\paragraph{Plot-to-Table}
In addition to relevant text on the screen, some visual document domains and types of problems can benefit from other types of intermediate structure.
An example is understanding charts and figures, whose underlying structured source data is not well captured by OCR systems.
Such structured source data is available in some datasets, \textit{e.g.}, \chartqa provides structured data tables extracted by ChartOCR~\cite{luochartocr2021}; but they can also be inferred for unannotated images through tools like DePlot~\cite{liu-2022-deplot}.

\paragraph{LLM-Programmer}
For problems involving numerical reasoning, we use a prompted LLM, \palm, to generate a simple program capturing common numerical reasoning patterns corresponding to user queries, given the question, answer, the full image OCR text and the structured table (see Figure~\ref{fig:rationale_generation} lower half for an example).
The programs are limited to the following formats: \texttt{Div(a,b)}; \texttt{Mul(a,b)}; \texttt{Avg(a list of numbers)}; \texttt{Sum(a list of numbers)}; \texttt{Diff(a,b)}; \texttt{Greater(a,b)}; \texttt{Less(a,b)}; \texttt{Find(str)}.
All programs except \texttt{Find(str)} have execution steps in the prompt templates, which explain how to connect the programs to arithmetic and comparison operations.
The last program type is applicable if numerical reasoning of the other types is not needed, and has no operation involved. Note that the program rationale is not executed by default, but is only used to guide the model towards the correct answer.

\paragraph{Multimodal-Verifier}
To determine the helpfulness of the rationale generated by other tools and the relevance of image augmentations, we employ a multi-task trained, large multimodal model \palix 55B~\cite{chen2023palix}.
We construct the text encoder input in the following format:
\begin{align*}
    \text{[rationale] Answer in en: [question]}
\end{align*}
The verifier takes in the image $I$ as input to the vision encoder, the question $q$ and the rationale $r$ as input to the text encoder. 
We use the log-probability of the gold answer (with and without conditioning on the rationale), and the correctness of the predicted answer (through greedy decoding), to define two measures of rationale helpfulness.

\subsection{Algorithm for rationale augmentation}
Here we list the detailed algorithm for rationale augmentation described in~\S\ref{subsec:augmentation}.

\begin{algorithm}[ht]
  \caption{Rationale Augmentation via Image Cropping}
  \label{algo:aug}
\begin{algorithmic}[1]
\State {\bfseries Input:} image $I$; question $q$; answer $a$; tools for rationale generation.
\State {\bfseries Output:} a set of cropped images, and a corresponding set of rationales.
\State Initialize the counter $j \leftarrow 0$, the cropped image set $\mathcal{I} \leftarrow \varnothing$ and the rationale set $\mathcal{R} \leftarrow \varnothing$.
\State Get the height $h$ and the width $w$ of the image $I$.
\If{$h\ge w$}
    \While{$wj<h$}
        \State $start \leftarrow wj / 2$
        \State $end \leftarrow \min(wj/2+w, h)$
        \State image $i_j \leftarrow$ crop $[start, end]$ on the height of $I$.
        \State Get rationales $r_j$ for $i_j, q, a$ from tools.
        \State $\mathcal{I} \leftarrow \mathcal{I} \cup i_j$; $\mathcal{R} \leftarrow \mathcal{R} \cup r_j$; $j \leftarrow j+1$.
    \EndWhile
\Else
    \While{$hj<w$}
        \State $start \leftarrow hj/2$
        \State $end \leftarrow \min(hj/2+h, w)$
        \State image $i_j \leftarrow$ crop $[start, end]$ on the width of $I$.
        \State Get rationales $r_j$ for $i_j, q, a$ from tools.
        \State $\mathcal{I} \leftarrow \mathcal{I} \cup i_j$; $\mathcal{R} \leftarrow \mathcal{R} \cup r_j$; $j \leftarrow j+1$.
    \EndWhile
\EndIf
\State \textbf{return} $\mathcal{I}, \mathcal{R}$
\end{algorithmic}
\end{algorithm}

\subsection{Student rationale generation for \tasktwo}
For student rationale generation, we cannot directly use the student trained on the whole training set, as it is likely to remember and replicate the tool-generated rationale but this would not be representative of its behavior on unseen data.

On \infovqa and \docvqa, we split the training data of into 3 folds. We train 3 student models, each takes in 2 folds as the train data and generates student rationale for the remaining fold.
On \chartqa, to avoid the distribution shift from the augmented set and human set, we split both augmented set and the human set into 3 holds, in total 6 folds. 
We train 6 student models, each takes in 5 folds for training and generates student rationale for the remaining fold.

Here, the student models are only trained to generate the question and the rationale, not the answer. 
The output format of the student models is
\begin{align*}
    \text{[question] \texttt{<s>} [rationale]}
\end{align*}
For each example, we sample 3 rationales to create the \tasktwo training set.

\subsection{Hyper-parameters}
Following the setup in~\citet{pix2struct}, for \pts-Base, we use an input sequence length of 6155 patches for \infovqa, and 4096 patches for \docvqa and \chartqa. 
We train with a batch size of 128 for \infovqa, and 256 for \docvqa and \chartqa, on 32 v3-Google Cloud TPUs.

For \pts-Large, we use an input sequence length of 3072 patches and train with a batch size of 64 for all datasets, on 64 v3-Google Cloud TPUs.

We train all the model with 10k steps, optimizing using Adafactor~\cite{shazeer2018adafactor}. The learning rate schedule uses a linear warmup of 1k steps to 0.01, followed by cosine decay to 0. On \infovqa and \docvqa, we select the model with the best ANLS score on the dev set for evaluation. On \chartqa, we select the model with the best RA on the dev augmented set for test evaluation.
We report all the results under a single-run setup.

\input{tables/datasets}
 
\subsection{Cost of external tool calls}
Many of the tools we used are publicly accessible, such as Google Cloud OCR\footnote{https://cloud.google.com/use-cases/ocr?hl=en} and PaLM-2 text bison\footnote{https://pypi.org/project/google-generativeai}. Based on the pricing, an estimate of the cost is approximately \$150 for OCR and \$400 for PaLM-2 to generate training data for student models.

\subsection{Scientific artifacts and licenses}
We evaluate on three public datasets, \infovqa, \docvqa and \chartqa, in our experiments.
\infovqa and \docvqa data is shared for non-commercial, research and educational purposes, which aligns with our use.
\chartqa is under GNU General Public License v3.0.
The questions in all three datasets are in English.
We put the statistics of our evaluated datasets in Table~\ref{tab:dataset}.

We finetune public models \pts and \matcha. They are under Apache License 2.0.

\section{Additional experimental analysis}
\label{appsec:analysis}

\subsection{Model ablations}
We show that \ourmethod also benefits stronger pretrained model such as~\cite{liu-2022-matcha}, while decoupling rationale and answer  prediction is harmful.

\input{tables/better_init_matcha}

\paragraph{What if we use a stronger pretrained model tailored to math reasoning as in \chartqa?}
We initialize our student model parameters with \matcha~\cite{liu-2022-matcha} instead of \pts before finetuning with \ourmethod on \chartqa (Table~\ref{tab:chartqa_matcha}).
\matcha is based on \pts-Base but has stronger numerical reasoning and other abilities obtained through additional pretraining on relevant data. 
We see that \ourmethod leads to consistent improvements over stronger \matcha models specialized for this domain.

\paragraph{Decoupling rationale and answer prediction.}
\ourmethod uses the same student model (with a single set of parameters $\theta$) to predict rationales and answers. 
In Figure~\ref{fig:student_quality}, ``Another Student'' refers to using a student model, with a separate set of parameters, only responsible for rationale prediction.
While training separate models for predicting different intermediate steps has been shown beneficial for ScienceQA~\cite{zhang2023multimodal}, this configuration results in slightly worse performance on \infovqa dev set. Moreover, it also adds engineering complexity, storage, and compute. 

\paragraph{Selecting appropriate rationales is important.}
Instead of using a simple customized program, we construct the rationale for \chartqa by structured table concatenated with text evidence.
The text evidence describes information in the figure that is relevant to the question and is predicted by \palm given the question, answer, structured table, and OCR, but does not specify a program that can be executed to obtain the answer.
For example, for the input in the lower half of Figure~\ref{fig:rationale_generation}, the text evidence generated by \palm in this setting is \textit{``No confidence value in 2017 is 5, confidence value in 2017 is 93''}.
The same \ourmethod training on evidence-based rationales achieves 83.4 / 33.0 RA on the \chartqa's augmented and human test sets, which is 1.4 / 5.0 points lower than the program-based rationales. 

\input{tables/other_approaches}

\subsection{Comparison to other approaches}
\label{appsubsec:other_approaches}
We make an additional comparison to other approaches, which may have different setups, such as the use of tools or LLMs at inference time, or the use of additional pretraining, in Table~\ref{tab:other_approaches}. We show that except the powerful pretrained model PaLI-3 (5B parameters), RD is better than other approaches under the setup of pixel-level image-to-text model without the use of external tools at inference time.

UniChart~\cite{masry2023unichart} is pretrained on chart-specific objectives, but on a larger corpus than \matcha. The pretraining data is augmented by knowledge distillation from LLMs.
Without further pretraining, \ourmethod shows better performance on \chartqa, initialized with \matcha.
DUBLIN~\cite{aggarwal2023dublin} proposes pretraining objectives at four different levels: language, image, document structure, and question-answering. It demonstrates high performance on \infovqa and \docvqa, at the cost of sacrificing the ability to understand charts.
In addition, UReader~\cite{ye2023ureader} designs a shape-adaptive cropping module to process high-resolution images. It is jointly finetuned on multiple VDU tasks with low-rank adaptation approach.
Cream~\cite{kim2023cream} utilizes contrastive learning to align the visual representation of the image and text representation of OCR and objects (generated from tools).
We show that \ourmethod is better than or close to Cream even under the setup where Cream uses tools in inference.  

UDOP~\cite{tang2022unifying} uses external OCR tool for text layout information at training and inference time. It is also pretrained on the IIT-CDIP scanned documents corpus, achieving great performance gains on \infovqa and \docvqa.

\input{tables/qualitative_tool_ocr}

\subsection{Qualitative analysis}
\label{appsubsec:qualitative}

We randomly select 5 examples in the dev set of \infovqa to illustrate that tool generated rationales extract relevant information from the visual context, which are helpful to answer the question (Table ~\ref{tab:qualitative_tool}).

We also randomly select 20 examples from the dev set of \infovqa for a qualitatively analysis of student generated rationales (Table ~\ref{tab:qualitative}). The first five examples are for the same inputs as the tool-generated rationale examples.
We observe that for 3 examples out of 5, the student generated rationales match the tool generated ones.
In the table, we list the student generated and TF-IDF extracted rationales, along with the question and the ground truth answer.
We compute the TF-IDF weight for each OCR block in the image, and measure the cosine similarity of the question to these OCR blocks.
Starting from the closest OCR block to the question, we gradually add more OCR blocks to the final TF-IDF string until it reaches 50 tokens under \pts tokenizer.
Note that this process is also applied to the TF-IDF and BERT embedding analysis in Figure~\ref{fig:student_quality}.

For more than 50\% of the student generated rationales, answers can be inferred from them without looking at the images. Also, 90\% of the student generated rationales are relevant to the answer.
It is possible for the student model to generate an irrelevant rationale, such as in the last row of Table~\ref{tab:qualitative}, the student rationale (\textit{27 \% fake or empty 28 \% inactive 43\% good}) is irrelevant to the question (\textit{Who uses the twitterid @Ev?}) as well as the answer (\textit{twitter co-founder evan williams}).
This observation verifies the importance of robustifying against student rationale errors during training.

\subsection{Validity of \palix as a verifier}
\label{appsubsec:verify_pali}

We choose \palix for filtering as it was the SOTA before PaLI-3 on our evaluated benchmarks (without OCR pipeline). We have verified \palix{}’s filtering ability on two axes.

\begin{figure}[ht]
    \centering
    \scalebox{.9}{
    \includegraphics[width=1\linewidth]{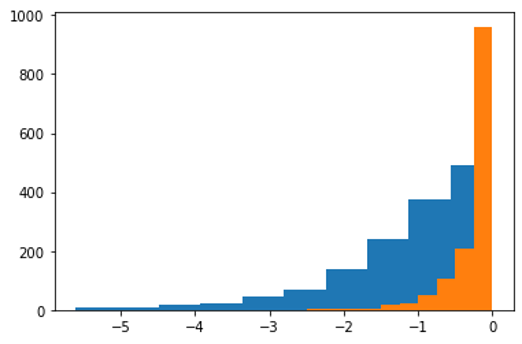}
    }
    \caption{The distribution of the log-likelihood of \palix prediction; $y$-axis is the number of examples. Orange bars show correct predictions and blue bars show the wrong predictions.}
    \label{fig:pali-dist}
\end{figure}

First, we checked if higher likelihood on \palix with rationales as the input implies the prediction is more likely to be correct. 
For the correct predictions, $\sim$75\% have log-likelihood greater than -0.25; for the wrong prediction, $\sim$70\% have log-likelihood lower than -0.6, which shows \palix is not over-confident on wrong predictions (see Figure~\ref{fig:pali-dist} for details).

Second, we evaluate on a small scale whether if the relevance filter is satisfied, the generated rationales are more useful.
We manually examined $\sim$5 examples and verified those rationales are actually useful. 
We give one example here where the \palix prediction is wrong without the rationale, but correct with the rationale.

\textbf{Question}: What was the name of 'Snapchat' when it was launched?

\textbf{Rationale}: Snapchat, an image messaging app, is launched in September after being tested as an IOS only app known as Picaboo in July. One of the founding members is pushed out before Snapchat is released., SHORT VIDEOS, MEGA VIEWS, Vine, a video hosting service specialising in short videos, goes public in June, Twitter acquires Vine for \$30 million in the October of that year.

\textbf{\palix's answer w/o rationale}: Snapchat

\textbf{\palix's answer w/ rationale}: Picaboo

\section{Extended related works}
\label{appsec:related}
Here we summarize related research in text only and visual language understanding, focusing on methods using intermediate reasoning steps.

\paragraph{Tool use in visual language understanding}
Using tools to augment the input in a prediction problem can be seen as using additional reasoning steps of specific type, \textit{i.e.}, calling a tool with a set of arguments and integrating its result with the rest of the context.
Much prior work on visual document understanding has relied on an OCR component~\cite{tang2022unifying, appalaraju2021docformer}. 

PaLI-X~\cite{chen2023palix} and the smaller PaLI-3 model~\cite{chen2023pali3}, which are image-and-text encoders paired with text decoders, achieve strong results both with and without additional OCR input. 
Since OCR extractions can be very long, \textit{e.g.}, \infovqa has images with OCR more than 1k tokens, the recognized text often needs to be truncated to a given maximal token length given pretrained model assumed token limits and efficiency considerations.
Other architectures are heavily centered on the recognized document text, with examples being TILT~\cite{tilt} and LayoutLM~\cite{layoutlmv3}.

In addition to OCR, de-plotting has been used as a pre-processing step to either augment or entirely replace the input image representation~\cite{liu-2022-deplot}. 
Both object detection and OCR are used as an auxiliary input by Cream~\cite{kim2023cream} to augment the vision feature.

Such works have not attempted to recognize text or structured data as an intermediate reasoning step using the same small model, as we have proposed in \ourmethod.

\paragraph{Tool use and chain-of-thought distillation}

Distilling text rationales from large teacher models has been shown successful by chain-of-thought distillation works~\cite{shridhar-etal-2023-distilling, li-etal-2023-symbolic, wang-etal-2023-scott} on NLP benchmarks, such as CommonsenceQA~\cite{talmor-etal-2019-commonsenseqa} and QuaRel~\cite{Tafjord2018QuaRelAD}.

MMCoT~\cite{zhang2023multimodal} and T-Sci~\cite{wang2023t} have utilized annotated or decomposed reasoning chains for improving vision-language reasoning on ScienceQA, which is not representative of the visual document understanding challenges we focus on (\textit{e.g.} text-only models can reach accuracy of over 79\% on this benchmark). In addition, these works only distill using our \taskone task, which we show is insufficient to teach the student model to produce high-quality rationales and be robust to potential errors. We also use a single small model instead of two different models for rationale and answer generation, reducing complexity and engineering cost, and focus on short rationales for efficiency. Finally, we use a broader set of tools instead of just one LLM chain-of-thought tool.

Toolformer~\cite{Schick2023ToolformerLM} trains smaller language models to call tools. Generic multimodal tool use solutions based on LLMs have also been proposed~\cite{gpt4tools}. However, these works do not replicate the results of tool output and replace them for efficiency.

\paragraph{Other related work on text-only models with intermediate reasoning steps}

Intermediate reasoning in text-only models has been successful through prompting large language models to perform a chain-of-thought~\cite{wei2023chainofthought}. 
More traditionally in NLP, smaller models have been shown to be able to successfully learn to generate semantic parses before predicting final answers, including when such parses are not directly annotated in training data~\cite{yih-etal-2016-value}.
Decomposing intermediate questions is also known to help small models on multistep text question answering~\cite{zhu2023chain}.
Marginalizing over multiple intermediate rationale possibilities has brought consistent gains~\cite{selfcons}.

The specific structure of reasoning chains (which can be guided by tailored prompting strategies for LLMs) used has been shown to have significant impact~\cite{zhou2023leasttomost, khot2023decomposed, Yao2023TreeOT}. 
In addition to text as intermediate predictions, generating programs has also been shown useful~\cite{chen2022program}.

\input{tables/flops}

\section{Detailed FLOPs analysis}
\label{appsec:flops}

We show that \ourmethod only increases the FLOPs of the Base model on \infovqa by around 1\%, those of the Large model by around 2\%, and uses around 10\% the FLOPs of the SOTA model, as listed in Table~\ref{tab:flops}.

We only consider the computation of transformer blocks of the encoder and the decoder, and ignore the small cost in the last linear layer for token generation. Most of the computation cost is from the attention and feed-forward layers, and we ignore the activation and normalization layers.
Notice that matrix multiplication of with dimension $[N,P]\times[P,M]$ uses FLOPs of $NM(2P-1)$; for simplicity, we use $2NMP$ to approximate.

For each self-attention layer, we suppose an input sequence length of $d_{q}$, a hidden size of $d_{h}$. 
The query, key, value matrix computation takes $6d_{q}d_{h}^2$, the multiplication of these three matrices takes $4d_{q}^2d_{h}$, and the linear transformation towards the output takes $2d_{q}d_{h}^2$. 
The total is $8d_{q}d_{h}^2+4d_{q}^2d_{h}$.

For each cross-attention layer, we suppose a query input sequence length of $d_{q}$, and a key-value input token sequence of $d_{k}$. 
The query, key, value matrix computation takes $2d_{q}d_{h}^2+4d_{k}d_{h}^2$, the multiplication of these three matrices takes $4d_{q}d_{k}d_{h}$, and the linear transformation towards the output takes $2d_{q}d_{h}^2$. 
The total is $4d_{q}d_{h}^2+4d_{k}d_{h}^2+4d_{q}d_{k}d_{h}$.

For one feed-forward layer, suppose the sequence length from the attention layer is $d_q$ and the hidden size from the attention layer is $d_h$ and the feed-forward size is $d_{f}$, the total computation is $6d_{q}d_{f}d_{h}$ if gated activation is used, otherwise $4d_{q}d_{f}d_{h}$.

Now we derive the formula of FLOPs for encoder-decoder models. 
We use $d_e$ and $d_d$ to denote the encoder sequence length, and the whole decoder sequence length, respectively. 
Given the models we discuss here all have same hidden dimension for the encoder and the decoder, we use $d_h$ to denote the hidden size and $d_{f}$ to denote the feed-forward size.
For simplicity, we assume a batch size of 1. 
The computation cost of each encoder layer, denoted with FCE, is
\begin{align*}
    \text{FCE}(d_{e}, d_{h}, d_{f})&=8d_{e}d_{h}^2+4d_e^{2}d_h+4d_{e}d_{f}d_{h}\\
    &+2[\![\text{Gated}]\!]d_{e}d_{f}d_{h},
\end{align*}
where $[\![\text{Gated}]\!]$ is the indicator function on whether the model uses gated activation.
Similarly, without caching the past attention matrices, the computation cost of each decoder layer, denoted with $\text{FCD}_\text{exact}$, is
\begin{align*}
&\text{FCD}_\text{exact}(d_{e}, d_{d}, d_{h}, d_{f})=4d_{e}d_{h}^2+\sum_{t=1}^{d_d}4d_{h}t^{2}\\&+(12d_h^{2}+4d_{e}d_{h}+4d_{f}d_{h}+2[\![\text{Gated}]\!]d_{f}d_{h})t.
\end{align*}
Notice the query, key, value matrices from the encoder output only have to be constructed once through the decoding time steps.

Instead, if we consider KV-caching and reusing the past attention matrices in the decoding~\cite{pope2023efficiently}, we can 
achieve the following at step $t$:
\begin{itemize}[nosep,noitemsep]
    \item reuse the first $t-1$ rows of the query, key, value matrices;
    \item reduce the matrix multiplication cost by a factor of $t$ with block matrix computation;
    \item for both self-attention and cross-attention, we only have to care about the last row of the output matrix.
\end{itemize}
Given the decoding with caching, we reduce the computation cost of each decoder layer to $\text{FCD}_\text{approx}$, written as
\begin{align*}
&\text{FCD}_\text{approx}(d_{e}, d_{d}, d_{h}, d_{f})=4d_{e}d_{h}^2+\sum_{t=1}^{d_d}4d_{h}t\\&+(12d_h^{2}+4d_{e}d_{h}+4d_{f}d_{h}+2[\![\text{Gated}]\!]d_{f}d_{h}).
\end{align*}
This formula matches the one provided by~\citet{Elbayad2020Depth-Adaptive}.
For a $N$-layer encoder-decoder model, the total computation cost is $N(\text{FCE}+\text{FCD}_\text{approx})$ and $N(\text{FCE}+\text{FCD}_\text{exact})$ with and without caching, respectively.

Based on the formula derived above, we start to compute FLOPs for specific models. Taking \infovqa as an example, the student generated rationales have $41.8$ tokens on average, the questions have $15.3$ tokens on average and the answers have $5.0$ tokens on average.

\paragraph{\pts-Base} The model has $N=12$, $d_h = 768$, $d_{f} = 2048$ and uses gated activation. For \infovqa, we have $d_e = 6155$, $d_d = 5$ for answer-only generation, and $d_d = 62$ for \ourmethod generation (including the question, rationale, and the answer).
Without caching, the total FLOPs computation is $2.63$E+$12$ for answer-only generation, and $3.46$E+$12$ for \ourmethod generation, resulting in a $\sim30\%$ increase of computation.
With caching, the total FLOPs computation is $2.62$E+$12$ for answer-only generation, and $2.65$E+$12$ for \ourmethod generation, resulting in a only $\sim1\%$ increase of computation.

\paragraph{\pts-Large} The model has $N=18$, $d_h = 1536$, $d_f = 3968$ and uses gated activation.  Similarly, for \infovqa, we have $d_e = 3072$, $d_d = 5$ for answer-only generation, and $d_d = 62$ for \ourmethod generation. With caching, the total FLOPs computation is $4.63$E+$12$ for answer-only generation, and $4.72$E+$12$ for \ourmethod generation, resulting in a only $\sim2\%$ increase of computation.

\paragraph{PaLI-3} We also estimate FLOPs for PaLI-3~\cite{chen2023pali3}, which is constructed by a 2B ViT-G/14 vision encoder and a 3B UL2 language encoder-decoder.

The vision encoder has $N=48$, $d_h=1536$, $d_f=8192$, and does not use gated activation. For evaluating on \infovqa, the model uses the resolution of 1064$\times$1064, which has $d_e=5776$ patches. The FCE formula gives the computation cost of $2.90$E+$13$.

The language encoder-decoder has $N=24$, $d_h=1024$, $d_f=16384$, and uses gated activation. 
We consider the extra text tokens ($15$ on average) from the question but not the ones from the OCR input.
Hence, we have $d_e\ge 5791$ and $d_d=5$.
With caching, the total computation cost of the UL2 language transformer is at least $1.91$E+$13$.

Combing two parts, the 5B PaLI-3 model uses FLOPs of at least $4.81$E+$13$ on the setup of the \infovqa task, which is 10 times more than \pts-Large with the \ourmethod generation.

\section{Prompt templates}
\label{appsec:prompt}
\begin{figure*}
    \centering
    \includegraphics[width=\textwidth]{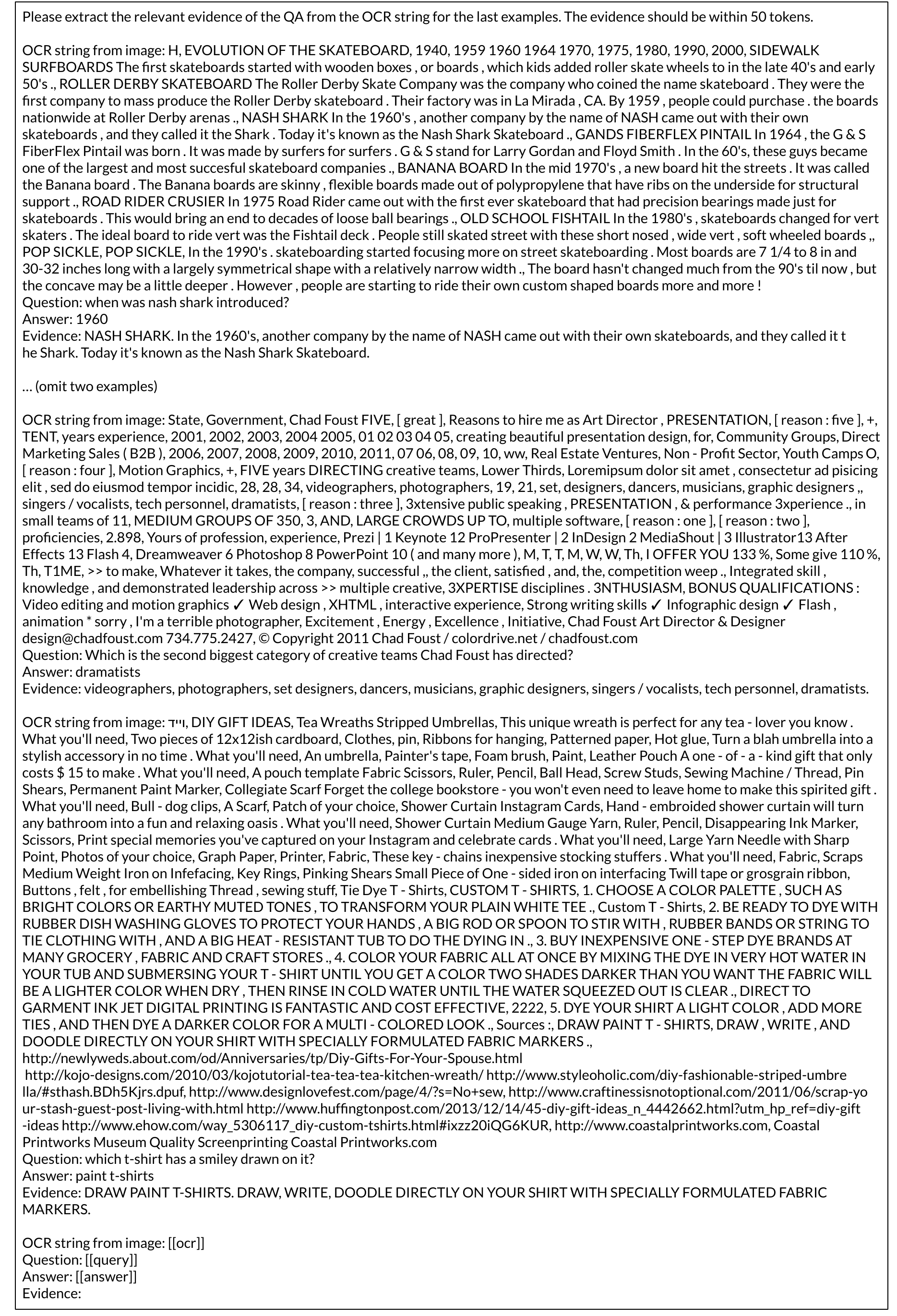}
    \caption{\infovqa prompt template.}
    \label{fig:infovqa_prompt}
\end{figure*}

\begin{figure*}
    \centering
    \includegraphics[width=\textwidth]{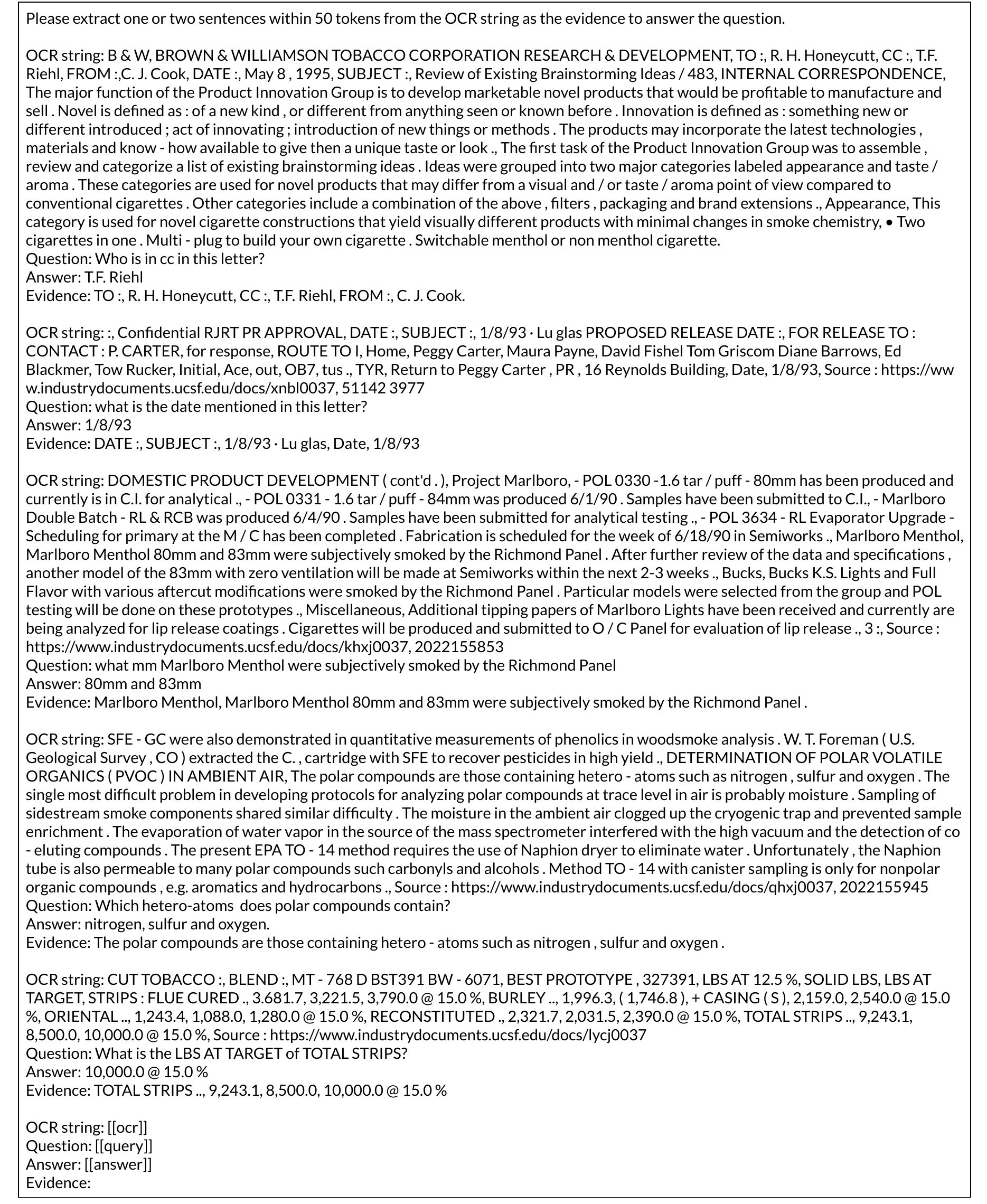}
    \caption{\docvqa prompt template.}
    \label{fig:docvqa_prompt}
\end{figure*}

\begin{figure*}
    \centering
    \includegraphics[width=0.97\textwidth]{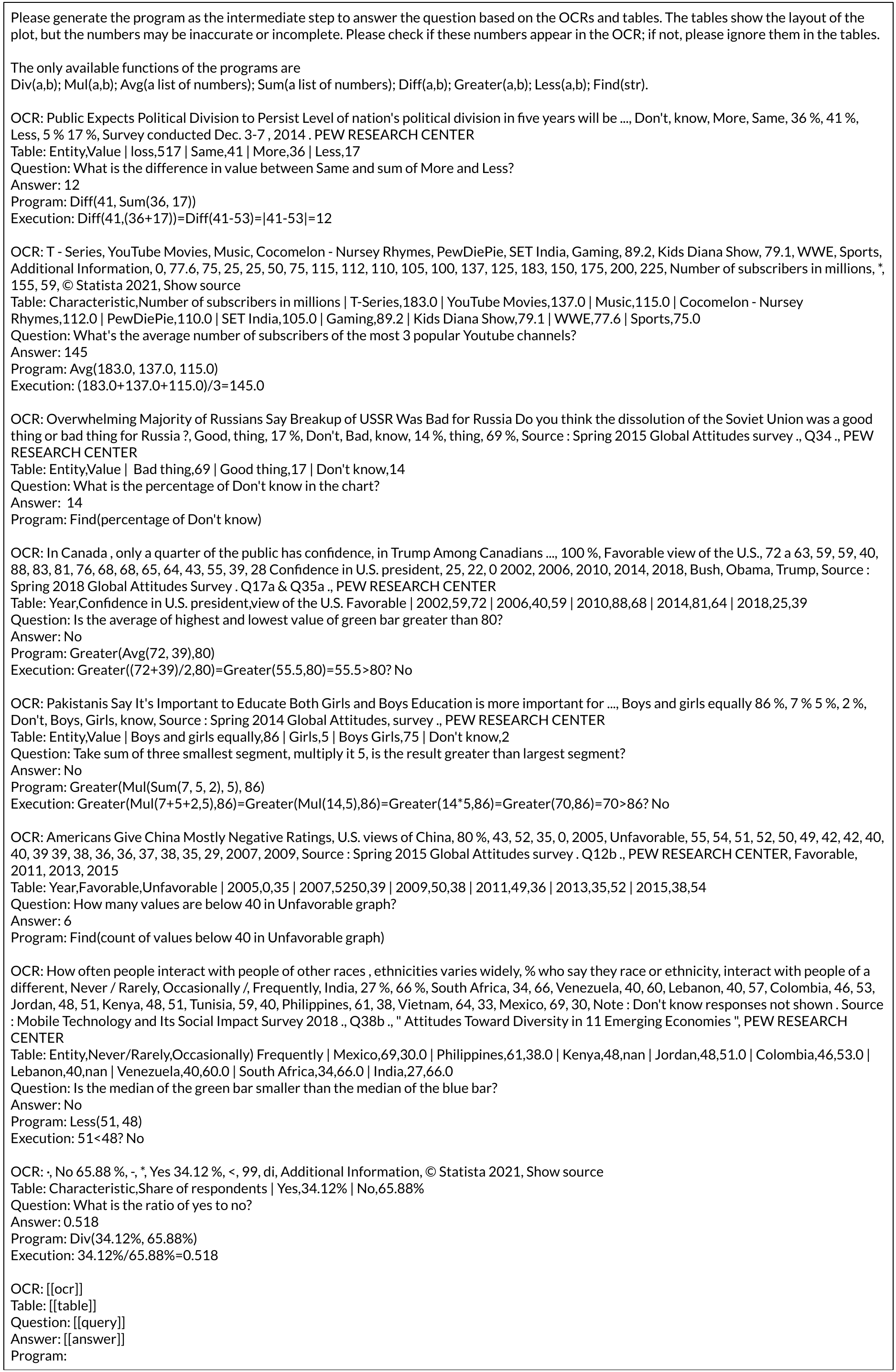}
    \caption{\chartqa prompt template.}
    \label{fig:chartqa_prompt}
\end{figure*}

We list the prompt templates for rationale generation on \infovqa, \docvqa and \chartqa in Fig.~\ref{fig:infovqa_prompt}, Fig.~\ref{fig:docvqa_prompt} and Fig.~\ref{fig:chartqa_prompt}, respectively.
The former two use 5-shot prompting for LLM-Summarizer and the last uses 8-shot prompting for LLM-Programmer. 

\input{tables/qualitative_ocr}

%% file: tables/datasets.tex
\begin{table}[t]
\setlength{\aboverulesep}{1pt}
\setlength{\belowrulesep}{1pt}
\centering
\small
\tabcolsep 5pt
\begin{tabular}{lcrrr}
    Dataset & Domain & Train & Dev & Test \\
    \toprule
    \infovqa & Documents & 23,946 & 2,801 & 3,288 \\
    \docvqa & Documents & 39,463 & 5,349 & 5,188 \\
    \chartqa-human &  \multirow{2}{*}{Illustrations} & 7,398 & 960 & 1,250 \\
    \chartqa-aug. & & 20,901 & 960 & 1,250 \\
    \bottomrule
\end{tabular}
\caption{Statistics of the datasets we evaluate on.}
\label{tab:dataset}
\end{table}

%% file: tables/better_init_matcha.tex
\begin{table}[t]
\setlength{\aboverulesep}{1pt}
\setlength{\belowrulesep}{1pt}
\centering
\small
\tabcolsep 6pt
\begin{tabular}{lrrrr}
    Method & \multicolumn{2}{c}{\chartqa Dev Set} & \multicolumn{2}{c}{\chartqa Test Set} \\
    & aug. & human & aug. & human \\
    \toprule
    \ansonly &  83.5 & 40.4 & 88.5 & 36.6 \\
    \qid &  84.6 & 40.2 & 89.7 & 37.5 \\
    \ourmethod &  \bf 86.0 & \bf 40.9 & \bf 90.8 & \bf 42.1 \\
    \bottomrule
\end{tabular}
\caption{We initialize the student model with \matcha, which has stronger numerical reasoning skills. \ourmethod also improves \matcha for \chartqa.}
\label{tab:chartqa_matcha}
\end{table}

%% file: tables/other_approaches.tex
\begin{table*}[t]
\setlength{\aboverulesep}{1pt}
\setlength{\belowrulesep}{1pt}
\centering
\small
\tabcolsep7pt
\begin{tabular}{lcccrrr}
    Model & Tool-use & Multi-dataset & Prompt LLM & \infovqa & \docvqa & \chartqa \\
    & in inference & fine-tuning & in inference \\
    \toprule
    Donut & \xmark & \xmark & \xmark & 21.7 & 67.5 & 41.8 \\
    \pts & \xmark & \xmark & \xmark & 40.0 & 76.6 & 59.5 \\
    \matcha & \xmark & \xmark & \xmark & 37.2 & 74.2 & 64.2 \\
    UniChart & \xmark & \xmark & \xmark & - & - & 66.3 \\
    DUBLIN & \xmark & \xmark & \xmark & 43.0 & \bf \color{blue} 80.7 & 35.2 \\
    UReader & \xmark & \cmark & \xmark & 42.2 & 65.4 & 59.3 \\
    Cream-Vicuna7B (w/o tools) & \xmark & \cmark & \cmark & 22.1 & 41.1 & 50.0 \\
    \bf \ourmethod (best model) & \xmark & \xmark & \xmark & \bf \color{blue} 44.3 & 79.0 & \bf \color{blue} 66.5 \\
    PaLI-3 (w/o OCR) & \xmark & \xmark & \xmark & \bf \color{red} 57.8 & \bf \color{red} 87.6 & \bf \color{red} 70.0 \\
    \midrule
    Cream-Vicuna7B (w/ tools) & \cmark & \cmark & \cmark & 43.5 & 79.5 & 63.0 \\    
    UDOP & \cmark & \xmark & \xmark & 47.4 & 84.7 & 60.7 \\
    PaLI-3 (w/ OCR) & \cmark & \xmark & \xmark & \bf 62.4 & \bf 88.6 & 69.5 \\
    DePlot & \cmark & \xmark & \cmark & - & - & \bf 79.3 \\
    \bottomrule
\end{tabular}
\caption{We compare the best model of \ourmethod (\pts-Large on \infovqa and \docvqa, \matcha on \chartqa) with other existing approaches, some of them (bottom part) have different setups. We show that except the powerful pretrained model PaLI-3, \ourmethod is better than other approaches under the same setup. {\color{red} Red} is the best model and {\color{blue} blue} is the second best.}
\label{tab:other_approaches}
\end{table*}

%% file: tables/qualitative_tool_ocr.tex
\begin{table*}[t]
\setlength{\aboverulesep}{1pt}
\setlength{\belowrulesep}{1pt}
\centering
\small
\tabcolsep 7pt
\begin{tabular}{p{4.8cm}p{7.4cm}p{2.3cm}}
Question & Tool Generated Rationales & GT Answer \\
\toprule
What is the cost of a cup of coffee in Luanda and Tokyo, taken together? & Cost of a Cup Of Coffee ( USD ), Cost of a Cup Of Coffee ( USD ), \$ 3.80, \$ 6.65, \$ 3.12, \$ 8.29, \$ & \$10.45 \\
\midrule
What are the points to be kept in mind while reading? & When you read you have to remember a lot of things, like: Characters Main plot Sub-plots. & characters, main plot, sub-plots \\
\midrule
What will the diastolic reading be if you have High blood pressure stage 2? & High Blood Pressure (Hypertension) Stage 2, 140 or higher, or, 90 or higher, Hypertensive Crisis, Higher than 180, (Call your doctor immediately), and/or, Higher than & 90 or higher \\
\midrule
Which country has the lowest count of critical care beds, China, India, or UK? & China, 3.6, India, 2.3. & india \\
\midrule
What is the meaning of the symbol "Hearts in Hearts" in Doodles? & Hearts in Hearts, Shy person. & shy person \\
\bottomrule
\end{tabular}
\caption{We show five randomly selected examples with tool generated rationales. The rationales are helpful to answer the question.}
\label{tab:qualitative_tool}
\end{table*}

%% file: tables/flops.tex
\begin{table}[t]
\setlength{\aboverulesep}{1pt}
\setlength{\belowrulesep}{1pt}
\centering
\small
\tabcolsep 16pt
\begin{tabular}{lc}
    Method & FLOPs \\
    \toprule
    \pts-Base, Ans-only & $2.62$E+$12$ \\
    \pts-Base, \ourmethod & $2.65$E+$12$ \\
    \midrule
    \pts-Large, Ans-only & $4.63$E+$12$ \\
    \pts-Large, \ourmethod & $4.72$E+$12$ \\
    \midrule
    PaLI-3, w/o OCR & $4.81$E+$13$ \\
    \bottomrule
\end{tabular}
\caption{FLOPs of evaluated approaches. \ourmethod only increase the FLOPs of Base model by around 1\%, Large model by 2\%, and uses less than 10\% the FLOPs of the SOTA model.}
\label{tab:flops}
\end{table}

%% file: tables/qualitative_ocr.tex
{
\centering
\onecolumn
\small
\begin{longtable}{p{2.7cm}p{4.8cm}p{4.8cm}p{1.5cm}}
Question & Student Generated & TF-IDF & GT Answer \\
\toprule
What is the cost of a cup of coffee in Luanda and Tokyo, taken together? & 1, LUANDA, Angola, 4,799,432, \$ 7,000, \$ 3.80, \$ 4.93, \$ 17.90, \$ 0.55, 2, TOKYO, Japan, 12,369,000, \$ 4.436, & Cost of Monthly, Cost of a, Rent Luxury Cup Of Coffee, 2 Bedroom ( USD ), ( USD ), This new found excess wealth has attracted many, Auxury retailers, hotels and restaurants to full indulgences & \$10.45 \\
\midrule
What are the points to be kept in mind while reading? & READING, LISTENING, DRINKING TEA, TO MUSIC, OR COFFEE, WALKING, 68 \%, 61 \%, 54 \%, 42 \%, 68 \%, 61 \%, 54 \%, 42 \%, Reduces stress, Just 6 minutes of reading :, 68 \%, 61 \%, 54 \% & READING, Alters your, state of mind, Reading develops theory of mind , which leads to better social connections ., Just 6 minutes of reading :, Why?, • It helps & characters, main plot, sub-plots \\
\midrule
What will the diastolic reading be if you have High blood pressure stage 2? & High Blood Pressure, 140 or higher, or, 90 or higher, 2 TRACK LEVELS, American Heart Association., Check. Change. Control., Health care providers can take blood pressure readings and provide recommendations. Check. Change. Control. helps you track your progress in reducing blood pressure., Check., Change., & High Blood Pressure ( Hypertension ) Stage 1, High Blood Pressure ( Hypertension ) Stage 2, Elevated Blood Pressure, BLOOD PRESSURE CATEGORY, & 90 or higher \\
\midrule
Which country has the lowest count of critical care beds, China, India, or UK? & China, 3.6, India, 2.3 & India O, China, The Countries With The Most Critical Care Beds Per Capita, Total number of critical care beds per 100,000 inhabitants in selected countries *, Sources : National & india \\
\midrule
What is the meaning of the symbol "Hearts in Hearts" in Doodles? & Hearth in Hearts Shy person & Hearts in Hearts, Shy person, Hearts Romantic person, Hearts with Arrow Daydreamer , idealistic person, What Do Your Doodles Mean?, Doodles can be a window into your inner thoughts & shy person \\
\midrule
What was the number of factory workers in the confederate states during the American Civil War? & 111K, 70K, 9K, 21K, 9K, 1.7K, 9K, Factories, Factory workers, Miles of Railroad, MILITARY & Factory workers, X, CIVIL WAR, BORDER STATES CONFEDERACY, Prisoners of War, African American :, Native, 178,975 American & 111k \\
\midrule
How many countries are hosting the 2015 ICC Cricket World Cup? & 2 COUNTRIES Australia and New Zealand - hosting the World Cup 2015 & COUNTRIES Australia and New Zealand - hosting the World Cup 2015, 3, Teams participating in the World Cup, ICC CRICKET WORLD CUP, 2015, AUSTRALIA & 2 \\
\midrule
Which of these countries is least corrupt - Great Britain, China or Mexico? & GREAT BRITAIN, \$ 37,500, RUSSIA \$ 18,000, MEXICO \$ 35,950, GREAT BRITAIN, \$ 37,500 & CHINA GREAT 2.6 \% BRITAIN, 2.5 \%, MEXICO, 35.9 \% CHINA S, CORRUPTION INDEX, ( OUT OF 100 & great britain \\
\midrule
How many points did Shaq score in 2000? & 49 \%, 47 \%, 47 \%, 13, 22 25 32 33 34 42 44 52, On Tuesday night, Shaquille O'Neal's number 34 will become the 9th retired number raised to the rafters at STAPLES Center. Here's a unique look at the intriguing & POINTS, POINTS, 2000/2001, 1999/2000, fff, 2001, 2002 & 2,344 \\
\midrule
How many countries have number of critical care beds less than 5? & United States, Germany, Italy, France, South Korea, Spain, Japan, United Kingdom, United States, 34.7, 29.2, 12.5, 11.6, 10.6, 9.7, Japan, 7.3, 6.6, 6.6, China, 3.6, India, 2.3 & The Countries With The Most Critical Care Beds Per Capita, Total number of critical care beds per 100,000 inhabitants in selected countries *, Sources : National Center for Biotechnology Information, Inten & 2 \\
\midrule
What percentage of women find video ads really annoying? & 80 \%, find video ads really annoying & 80 \%, find video ads really annoying, \% women who watch online video, Majority of women watch online video in the afternoon or evening, 47 \% watch video for up to 10 minutes a & 80\% \\
\midrule
In 2009, how many pedestrian men died? & In 2009, 157 Pedestrian Deaths, \url{http://www.nj.gov/njsp/info/fatalace/2009\_fatal\_crash.pdf}, MALE :, 112, FEMALE :, 45, MALEP: 45, & Pedestrian Deaths in Southern New Jersey Look Both Ways Before You Cross, In 2009 , 157 Pedestrian Deaths, Between 2007 and 2009 the highest & 112 \\
\midrule
What percentage of clothing and consumer electronic products of men photographed by mobile shoppers, taken together? & 22 \%, 22 \%, 32 \%, 18 \%, 4 \%, 5 \%, 13 \%, 2 \%, 20 \%, 30 \%, PRODUCTS PHOTOGRAPHED BY MOBILE SHOPPERS, 15 \%, At work, 25 \%, 12 \%, In the & PRODUCTS PHOTOGRAPHED BY MOBILE SHOPPERS, Consumer Clothing electronics, MEN, WHERE MEN AND WOMEN DO, THEIR MOBILE SHOPPING, TYPES & 44\% \\
\midrule
What is the value of New York Knicks? & NEW YORK KNICKS \$ 3.30B & NEW YORK KNICKS \$ 3.30B, NEW YORK METS \$ 2.00B, NEW YORK GIANTS \$ 3.10B, NEW YORK YANKE & \$3.30b \\
\midrule
How much more is the value of Barcelona FC when compared to Real Madrid (\$bn)? & BARCELONA FC \$ 3.64B, NEW YORK KNICKS \$ 3.30B, LOS ANGELES LAKERS \$ 3.00B, CHICAGO BULLS \$ 2.60B, GOLDEN STATE WARRIORS \$ 2.60B, CHICAGO BULLS \$ 2.50B, BRO & REAL MADRID, \$ 3.58B, BARCELONA FC, \$ 3.64B, A mountain of sponsorship and advertising cash keeps Man U king of the soccer castle, though Barcelona & 0.06 \\
\midrule
Which is the second last tip for staying healthy? & Don't touch your, face, Avoid close contact with someone who's, sick, Clean and disinfect surfaces and objects people frequently touch & Tips for staying healthy, ON, What to do if you feel sick, Stay home, Most people with COVID - 19 have mild to moderate symptoms and can recover at home. Rest up and prevent germs from spreading by staying home & wear a cloth face mask in public \\
\midrule
What percent of adults in age group 65+, buy their food based on the 'availability of nutritious food'? & 33 \%, 28 \%, 21 \%, 32 \%, 11 \%, 17 \%, 15 \%, Making it easier for the 50+ to eat more nutritious foods, i, 56 \%, Help find information on fruits \& vegetables, Source : AARP Foundation : Food Insecurity & Food Availability, AARP®<unk>, FOUNDATION, A recent AARP Foundation survey of 1,000 low - income adults age 50+ reveals that, in the past 12 months, two in & 15\% \\
\midrule
Who provides statements for the presentencing investigation report? & ANALYSIS OF LEGAL HISTORY, ANALYSIS OF LEGAL HISTORY, OI, Snapshot of the DV Criminal History including, Domestic Incident Report ( DIR ) history, How many arrests in DV related crimes? Convictions?, • Stalking history, • Protective orders?, • Level of compliance if under supervision before?, • Current release status, • Jail days credited, Domestic Incident Report ( DIR ) history, • & THINGS TO INCLUDE WHEN CREATING A PRESENTENCING INVESTIGATION REPORT, • Arrest Report / DIR • Depositions Summary of Witness Statements, Review Police report & arresting officer, victim \\
\midrule
What happened first; Gaza conflict or Scottish independence? & GAZA CONFLICT August 1 : 64K Peak Shares & GAZA CONFLICT August 1 84K Peak Shares SCOTTISH INDEPENDENCE September 14 35K Peak Shares, CRIMEAN INDEPENDENCE March 17 & gaza conflict \\
\midrule
Who uses the twitter id @Ev? & 27 \% fake or empty 28 \% inactive 43 \% good & Twitter co - founder Evan Williams @Ev, WHOLESALERS, IN DARK CORNERS OF THE INTERNET , THEY PLY TOOLS TO OVERRIDE TWITTER'S RULES, THE & twitter co-founder evan williams \\
\bottomrule
\caption{We show 20 random selected examples with student generated or TF-IDF extracted rationales. The first 5 examples are the same as in Table~\ref{tab:qualitative_tool}, where 60\% of student generated rationales match the tool generated ones. For more than 50\% of the student generated rationales, answers can be inferred from them without looking at the images. 90\% of the student generated rationales are relevant to the answer, others are irrelevant.}
\label{tab:qualitative}
\end{longtable}

\twocolumn
}